%% file: relativistic.tex
\documentclass{article}

\usepackage[utf8]{inputenc} % allow utf-8 input
\usepackage{graphicx}
\usepackage{caption}
\usepackage{subcaption}
\usepackage[T1]{fontenc}    % use 8-bit T1 fonts
\usepackage{hyperref}       % hyperlinks
\usepackage{url}            % simple URL typesetting
\usepackage{booktabs}       % professional-quality tables
\usepackage{amsfonts}       % blackboard math symbols
\usepackage{nicefrac}       % compact symbols for 1/2, etc.
\usepackage{microtype}      % microtypography
\usepackage{todonotes}
\usepackage{amsmath}
\usepackage{wrapfig}
\usepackage[left=3cm,right=3cm,top=3cm,bottom=3cm]{geometry}
\title{Relativistic Monte Carlo}

\author{Xiaoyu Lu\thanks{These authors contributed equally.}\\
  Department of Statistics\\
  University of Oxford\\
  \and
  Valerio Perrone\footnotemark[1] \\
  Department of Statistics \\
  University of Warwick \\
  \and
  Leonard Hasenclever \\
  Department of Statistics\\
  University of Oxford\\
  \and
  Yee Whye Teh \\
  Department of Statistics\\
  University of Oxford\\
  \and
  Sebastian J. Vollmer \\
  Department of Statistics\\
  University of Oxford}

\def\RR{\mathbb{R}}

\begin{document}

\maketitle

\begin{abstract}
  \input{abstract}

\end{abstract}

\input{introduction}

\input{relativistichmc}
\input{sgrhmc}\label{sec:sgld}
\input{thermostat}

\section{Experiments}\label{sec:experiments}
\subsection{Small examples}

\input{toy_example}

\input{LR}

\subsection{Neural Network}
\input{MNIST}
\input{conclusion}

\input{ack}

\small
\bibliographystyle{unsrt}
\bibliography{refs}
\end{document}

%% file: abstract.tex
% !TEX root = nips_paper_main.tex

Hamiltonian Monte Carlo (HMC) is a popular Markov chain Monte Carlo (MCMC) algorithm that generates proposals for a Metropolis-Hastings algorithm by simulating the dynamics of a Hamiltonian system. However, HMC is sensitive to large time discretizations and performs poorly if there is a mismatch between the spatial geometry of the target distribution and the scales of the momentum distribution. In particular the mass matrix of HMC is hard to tune well.

 In order to alleviate these problems we propose relativistic Hamiltonian Monte Carlo, a version of HMC based on relativistic dynamics that introduce a maximum velocity on particles.  We also derive stochastic gradient versions of the algorithm and show that the resulting algorithms bear interesting relationships to gradient clipping, RMSprop, Adagrad and Adam, popular optimisation methods in deep learning.  Based on this, we develop relativistic stochastic gradient descent by taking the zero-temperature limit of relativistic stochastic gradient Hamiltonian Monte Carlo.  In experiments we show that the relativistic algorithms perform better than classical Newtonian variants and Adam.

%% file: introduction.tex
\section{Introduction}

Markov chain Monte Carlo (MCMC) techniques based on continuous-time physical systems allow the efficient simulation of posterior distributions, and are an important mainstay of Bayesian machine learning and statistics.  Hamiltonian Monte Carlo (HMC) \cite{duane1987hybrid,Neal,Stan,NUTS} is based on Newtonian dynamics on a frictionless surface, and has been argued to be more efficient than techniques based on diffusions \cite{MALA}.  On the other hand, stochastic gradient MCMC techniques based on diffusive dynamics \cite{Welling2011,Ma2015,Ding2014,Chen2014} have allowed scalable Bayesian learning using mini-batches.

An important consideration when designing such MCMC algorithms is adaptation or tuning to the geometry of the space under consideration \cite{GirolamiCalderhead,BeskosRoberts,PattersonTeh}.  To give a concrete example, consider HMC.  Let $f(\theta)$ be a target density which can be written as $f(\theta)\propto e^{-U(\theta)}$ where $U(\theta)$ is interpreted as the potential energy of a particle in location $\theta$.  HMC introduces an auxiliary momentum variable $p$ so that the joint distribution is $f(\theta,p)\propto e^{-H(\theta,p)}$ where the Hamiltonian is $H(\theta,p) = U(\theta)+\frac{1}{2m}p^\top p$. The quantity $\frac{1}{2m}p^\top p$, where $m$ is the mass of the particle, represents the kinetic energy. Denoting by $\dot{\theta}$ and $\dot{p}$ the time derivative of $\theta$ and $p$, the leapfrog discretisation \cite{Neal} of Hamilton's equations $\dot{\theta} = \frac{\partial H}{\partial p}$ and $\dot{p} = -\frac{\partial H}{\partial \theta}$ gives
\begin{align}
p_{t+1/2} &\leftarrow p_t -\textstyle \frac{1}{2}\epsilon\nabla U(\theta_t),  &
\theta_{t+1} &\leftarrow \theta_t + \textstyle \epsilon \frac{p_{t+1/2}}{m}, &
p_{t+1} &\leftarrow p_{t+1/2} -\textstyle \frac{1}{2}\epsilon\nabla U(\theta_{t+1})
\end{align}
where $\epsilon$ is the time discretisation and the velocity is $\frac{p_{t+1/2}}{m}$.  If $m$ is too small, the particle travels too fast leading to an accumulation of discretisation error.  To compensate, $\epsilon$ needs to be set small and the computational cost required increases.  On the other hand, if $m$ is too large, the particle travels slowly resulting in slow mixing of the resulting Markov chain.  While the mass parameter can be tuned, e.g. to optimise acceptance rate according to theory \cite{BeskosRoberts}, it only incidentally controls the velocity which ultimately affects the discretisation error and algorithm stability.

In this paper, we are interested in making MCMC algorithms based on physical simulations more robust by directly controlling the velocity of the particle.  This is achieved by replacing Newtonian dynamics in HMC with relativistic dynamics \cite{Einstein}, whereby particles cannot travel faster than the ``speed of light''.  We also develop relativistic variants of stochastic gradient MCMC algorithms and show that they work better and are more robust than the classical Newtonian variants.  

The relativistic MCMC algorithms we develop have interesting relationships with a number of optimisation algorithms popular in deep learning.  Firstly, the maximum allowable velocity (speed of light) is reminiscent of gradient clipping \cite{PascanuMikolovBengio}.  Our framework gives Bayesian alternatives to gradient clipping, in the sense that our algorithms demonstrably sample from instead of optimising the target distribution (exactly or approximately).  Secondly, the resulting formulas (see \eqref{reldynamics}), which include normalisations by $L_2$ norms,  bear strong resemblances to (but are distinct from) RMSprop, Adagrad and Adam \cite{Hinton,DuchiSinger,BaKingma}.  Motivated by these connections, we develop a relativistic stochastic gradient descent (SGD) algorithm by taking the zero-temperature limit of relativistic SGHMC, and show in an experiment on feedforward networks trained on MNIST that it achieves better performance than Adam.

%% file: relativistichmc.tex
\section{Relativistic Hamiltonian Dynamics}

Our starting point is the Hamiltonian which governs dynamics in special relativity \cite{Einstein},
\begin{align}
H(\theta,p) &= U(\theta) + K(p) \\
K(p) &= mc^2\left(\frac{p^\top p}{m^2c^2}+1\right)^{\frac{1}{2}}
\label{relhamiltonian}
\end{align}
where the target density is $f(\theta) \propto e^{-U(\theta)}$, for $\theta\in \RR^d$ interpreted as the 
position of the particle, $p\in\RR^d$ is a momentum variable, and $K(p)$ is the relativistic kinetic energy.  The two 
tunable hyperparameters are a scalar ``rest mass'' $m$ and the ``speed of light'' $c$ which bounds the particle's speed.  The joint distribution  $f(\theta,p)\propto 
e^{H(\theta,p)}$ is separable, with the momentum variable having marginal distribution $\propto e^{-K(p)}$, a multivariate generalisation of the symmetric hyperbolic distribution.

The resulting dynamics are given by Hamilton's equations, which read
\begin{align}
\dot{\theta}  &= \frac{\partial H}{\partial p} = M^{-1}(p) p, &
M(p) &= m\left(\frac{p^\top p}{m^2c^2} + 1\right)^{\frac{1}{2}} \nonumber \\
\dot{p} &= -\frac{\partial H}{\partial \theta} = -\nabla U(\theta),
\label{reldynamics}
\end{align}
where $M(p)$ can be interpreted as a relativistic mass and $M^{-1}(p) p $ is the velocity of the particle (c.f.\ the velocity under Newtonian dynamics is $m^{-1} p$).  Note that the relativistic mass is lower bounded by and increases asymptotically to $\|p\|/c$ as the momentum increases, so that the speed $M^{-1}(p) \|p\|$ is upper bounded by  and asymptototes to $c$.
On the other hand, the larger the rest mass $m$ the smaller the typical ``cruising'' speed of the particle is.  Conversely, as $m\rightarrow 0$ the particle will travel at the speed of light at all times, i.e.\ it behaves like a photon.  This gives an intuition for tuning both hyperparameters $c$ and $m$ based on knowledge about the length scale of the target density: we  choose $c$ as an upper bound on the speed at which the parameter of interest $\theta$ changes at each iteration, while we choose $m$ to control the typical sensible speed at which the parameter changes.  We will demonstrate this intuition in the experimental Section \ref{sec:experiments}.

In very high dimensional problems (e.g.\ those in deep learning, collaborative filtering or probabilistic modelling), the maximum overall speed imposed on the system might need to be very large so that reasonably large changes in each coordinate are possible at each step of the algorithm.  This means that each coordinate could in principle achieve a much higher speed than desirable.  An alternative approach is to upper bound the speed at which each coordinate changes by choosing the following relativistic kinetic energy
\begin{align}
K(p) &= \sum_{j=1}^d m_jc_j^2\left(\frac{p_j^2}{m_j^2c_j^2}+1\right)^{\frac{1}{2}},
\label{eq:separable}
\end{align}
where $j$ indexes the coordinates of the $d$-dimensional system, and each coordinate can have its own mass $m_j$ and speed of light $c_j$.  This leads to the same Hamiltonian dynamics \eqref{reldynamics}, but with all variables interpreted as vectors, and all arithmetic operations interpreted as element-wise operations. Experimental results will be based on the separable variant which showed consistently better performance. For simplicity, in the theoretical sections we will describe only the non-separable version \eqref{relhamiltonian}. 

\subsection{Relativistic Hamiltonian Monte Carlo}

As a demonstration of the relativistic Monte Carlo framework, we derive a relativistic variant of the Hamiltonian Monte Carlo (HMC) algorithm \cite{Neal,duane1987hybrid}.  In the following, we will refer to all classical variants as Newtonian as they follow Newtonian dynamics (e.g.\ \emph{Newtonian HMC (NHMC)} vs \emph{relativistic HMC (RHMC)}).  

Each iteration of HMC involves first sampling the momentum variable, followed by a series of $L$ leapfrog steps, followed by a Metropolis-Hastings accept/reject step.  The momentum can be simulated by first simulating the speed $\|p\|$ followed by simulating $p$ uniformly distribution on the sphere with radius $\|p\|$.  The speed $\|p\|$ has marginal distribution given by a symmetric hyperbolic distribution, for which specialised random variate generators exist.  Alternatively, the density is log-concave, and we used adaptive rejection sampling to simulate it.
The leapfrog steps \cite{calvo1994numerical} with stepsize $\epsilon$ follows  \eqref{reldynamics} directly: Set $\theta_0,p_0$ to the current location and momentum and for $t=1,\ldots, L$,
\begin{align}
p_{t+1/2} &\leftarrow p_t -\textstyle \frac{1}{2}\epsilon\nabla U(\theta_t) \nonumber \\
\theta_{t+1} &\leftarrow \theta_t + \epsilon M^{-1}(p_{t+1/2}) p_{t+1/2} \nonumber \\
p_{t+1} &\leftarrow p_{t+1/2} - \textstyle  \frac{1}{2}\epsilon\nabla U(\theta_{t+1})
\end{align}
The leapfrog steps leave the Hamiltonian $H$ approximately invariant and is volume-preserving \cite{Leimkuhler2016}, so that the MH acceptance probability is simply $\min(1,\exp(-H(\theta_L,p_L)+H(\theta_0,p_0)))$.

Observe that the momentum $p$ is unbounded and may become very large in the presence of large gradients in the potential energy.  However, the size of the $\theta$ update is bounded by $\epsilon c$ and therefore the stability of the proposed sampler can be controlled. This behaviour is essential for good algorithmic performance on complex models such as neural networks, where the scales of gradients can vary significantly across different parameters and may not be indicative of the optimal scales of parameter changes.  This is consistent with past experiences optimising neural networks, where it is important to adapt the learning rates individually for each parameter so that typical parameter changes stay in a sensible range \cite{PascanuMikolovBengio,Hinton,DuchiSinger,BaKingma,csimcsekli2016stochastic}.    Such adaptation techniques have also been explored for stochastic gradient MCMC techniques \cite{preconditioned,distbayes}, but we will argue in Sections \ref{sec:sgld} and \ref{sec:experiments} that they introduce another form of instability that is not present in the relativistic approach.

%% file: sgrhmc.tex
\section{Relativistic Stochastic Gradient Markov Chain Monte Carlo}

In recent years stochastic gradient MCMC (SGMCMC) algorithms have been very well explored as methods to scale up Bayesian learning by using mini-batches of data \cite{Welling2011,Chen2014,Ding2014,Ma2015,Shang2015}.  In this section we develop relativistic variants of SGHMC \cite{Chen2014} and SGNHT \cite{Ding2014,Shang2015}.  These algorithms include momenta, which serve as reservoirs of previous gradient computations, thus can integrate and smooth out gradient signals from previous mini-batches of data.  As noted earlier, because the momentum can be large, particularly as the stochastic gradients can have large variance, the resulting updates to $\theta$ can be overly large, and small values of the step size are required for stability, leading potentially to slower convergence.  This motivates our development of relativistic variants.

We make use of the framework of \cite{Ma2015} for deriving SGMCMC algorithms. However, we like to note the same characterisations have already been obtained much earlier in \cite{villani2009hypocoercivity,pavliotis2014stochastic} and partial results even much earlier in the physics literature. Let $z$ be a collection of variables with target distribution $f(z) \propto e^{-H(z)}$.  Consider an SDE in the form
\begin{align}
dz &= - [D(z)+Q(z)] \nabla H(z) dt + \Gamma(z) dt + \sqrt{2 D(z)} dW &
\Gamma_i(z) &= \sum_{j=1}^d  \textstyle
\frac{\partial [D_{ij}(z) + Q_{ij}(z)]}{\partial z_j} 
\label{eq:Ma}
\end{align}
where $D(z)$ is a symmetric positive-definite diffusion matrix, $Q(z)$ is a skew-symmetric matrix which describes energy-conserving dynamics, $\Gamma(z)$ is a correction factor, and $W$ is the $d$-dimensional Wiener process (Brownian motion).  \cite{Ma2015} showed that under mild conditions the SDE converges to the desired stationary distribution $f(z)$.  Hence in the following we simply have to choose the appropriate $z$, $D$ and $Q$.  Once the correction factor $\Gamma$ is computed, the SDE discretised, and a stochastic estimate $\nabla \tilde{U}(z)$ for $\nabla U(z)$ substituted, we obtain a correct relativistic SGMCMC algorithm.  The stochastic gradient has  asymptotically negligible variance compared to the noise injected by $W$.

\subsection{Relativistic Stochastic Gradient Hamiltonian Monte Carlo}

Suppose our noisy gradient estimate $\nabla \tilde{U}(\theta)$ of $\nabla U(\theta)$ is based on a minibatch of data. Then, appealing to the central limit theorem, we can assume that $\nabla \tilde{U}(\theta) \approx \nabla U(\theta) + \mathcal{N}(0,B(\theta))$. Let $z=(\theta, p)$ and $H(z)$ be the relativistic Hamiltonian in \eqref{relhamiltonian}. Choosing 
\begin{align}
D(z) = \left(\begin{array}{cc} 0 & 0\\ 0 & D\end{array}\right),
Q(z) = \left(\begin{array}{cc} 0 & -I\\ I & 0\end{array}\right),
\text{and thus } \Gamma(z) = \mathbf{0},
\end{align}
where $D$ is a fixed symmetric diffusion matrix results in the following relativistic SGHMC dynamics:
\begin{center}
$\begin{pmatrix} 
d \theta \\ 
dp
\end{pmatrix}  = 
\begin{pmatrix} 
M^{-1}(p)p \\ 
-\nabla U(\theta)-DM^{-1}(p)p
\end{pmatrix}  dt + 
\begin{pmatrix} 
0 & 0\\ 
0 & \sqrt{2D}
\end{pmatrix} dW_t
$\end{center}
Using a simple Euler-Maruyama discretisation, the relativistic SGHMC algorithm is,
\begin{align}
p_{t+1} &\leftarrow p_t - \epsilon_t \nabla \tilde{U}(\theta_t) -\epsilon_t D M^{-1}(p_t)p_t
+\mathcal{N}(0,\epsilon_t(2D-\epsilon_t \hat{B}_t)) \nonumber \\
\theta_{t+1} &\leftarrow \theta_t + \epsilon_t M^{-1}(p_{t+1})p_{t+1}
\end{align}
where $\hat{B}$ is an estimate of the noise coming from the stochastic gradient $B(\theta)$. 
The term $DM^{-1}(p)p$ can be interpreted as friction, which prevents the kinetic energy to build up and corrects for the noise coming from the stochastic gradient. 

It is useful to compare RSGHMC with preconditioned SGLD \cite{preconditioned,distbayes} which attempt to adapt the SGLD algorithm to the geometry of the space, using adaptations similar to RMSProp, Adagrad or Adam.  The relevant term above is the update $M^{-1}(p_{t+1})p_{t+1}$ to $\theta_{t+1}$:
\begin{align}
M^{-1}(p_{t+1})p_{t+1} = \frac{p_{t+1}}{\sqrt{\frac{p_{t+1}^\top p_{t+1}}{c^2} + m^2}}
\end{align}
Note the surprising similarity to RMSProp, Adagrad and Adam, with the main difference being that the relativistic mass adaptation uses the current momentum instead of being separately estimated using the square of the gradient.  This has the advantage that the relativistic SGHMC enforces a maximum speed of change. In contrast, preconditioned SGLD has the following failure mode which we observe in Section \ref{sec:experiments}: when the gradient is small, the adaptation scales up the gradient so that the gradient update has a reasonable size.  However it also scales up the injected noise, which can end up being significantly larger than the gradient update, and making the algorithm unstable.

\subsection{Relativistic Stochastic Gradient Descent (with Momentum)}
\label{sec:RSGD}

Motivated by the relationship to RMSprop, Adagrad and Adam, we develop a relativistic stochastic gradient descent (RSGD) algorithm with momentum by taking the zero-temperature limit of the RSGHMC dynamics. This idea connects to Santa \cite{chen2015bridging}, a recently developed algorithm where an annealing scheme on the system temperature makes it possible to obtain a stochastic optimization algorithm starting from a Bayesian one. 

From thermodynamics \cite{laurendeau2005statistical}, the canonical (Gibbs Boltzmann) density is  proportional to $e^{-\beta U(z)}$ where $\beta = 1/k_BT$, $k_B$ begin the Boltzmann constant and $T$ the temperature. Previously we have been using $\beta=1$ which corresponds to the posterior distribution.  For general $\beta$, 
\begin{align}
\begin{pmatrix} 
d \theta \\ dp
\end{pmatrix}  = 
\begin{pmatrix} 
\beta M^{-1}(p)p \\ 
\beta\left(-\nabla U(\theta)-DM^{-1}(p)p\right)
\end{pmatrix}  dt + 
\begin{pmatrix} 
0 & 0\\ 
0 & \sqrt{2D}
\end{pmatrix} dW
\end{align}
By taking $\beta\rightarrow \infty$ the target distribution becomes more peaked around the MAP estimator.  Simulated annealing \cite{geman1986diffusions,andrieu2000simulated,chen2015bridging}, which increases $\beta\rightarrow\infty$ over time, forces the sampler to converge to a MAP estimator. Instead, we can derive RSGD by rescaling time as well, guaranteeing a non-degenerate limit process. Letting $\hat{\theta}(t) = \theta(\beta t)$, $\hat{p}(t) = p(\beta t)$, so that
\begin{align}
\begin{pmatrix} 
d \hat{\theta} \\ d\hat{p}
\end{pmatrix}  = 
\begin{pmatrix} 
 M^{-1}(\hat{p})\hat{p} \\ 
-\nabla U(\hat{\theta})-DM^{-1}(\hat{p})\hat{p}
\end{pmatrix}  dt + 
\begin{pmatrix} 
0 & 0\\ 
0 & \sqrt{\frac{2D}{\beta}}
\end{pmatrix} dW
\end{align}
and letting $\beta\rightarrow\infty$, we obtain the following ODE:
\begin{align}
\begin{pmatrix} 
d \theta \\ dp
\end{pmatrix}  = 
\begin{pmatrix} 
 M^{-1}(p)p \\ 
-\nabla U(\theta)-DM^{-1}(p)p
\end{pmatrix}  dt 
\end{align}
Discretising the above then gives RSGD. Notice that if the above converges, i.e.\ $\dot{\theta}=\dot{p}=0$, it does so at a critical point of $U$.  Similar to other adaptation schemes, RSGD adaptively rescales the learning rates for different parameters, which enables effective learning especially in high dimensional settings. Moreover, the update in each iteration is upper bounded by the speed of light. Our algorithm differs from others through the use of a momentum, and adapting based on the momentum instead of the average of squared gradients.

%% file: thermostat.tex
\section{A Stochastic Gradient Nosé-Hoover Thermostat for Relativistic Hamiltonian Monte Carlo}
Borrowing a second concept from physics, SGHMC can be improved by introducing a dynamic variable $\xi$ that adaptively increases or decreases the momenta. The new variable $\xi$ can be thought of as a \emph{thermostat} in a statistical physics setting and its dynamics expressed as
\begin{equation}
	d\xi = \frac{1}{d}\left(p^Tp -d\right)dt.
\end{equation}
The idea is that the system adaptively changes the friction for the momentum, `heating' or `cooling down' the system. The dynamics of this new variable, known as Nosé-Hoover \cite{nhthermo} thermostat due to its links to statistical physics, has been shown to be able to remove the additional bias due to the stochastic gradient provided that the noise is isotropic Gaussian and spatially constant (\cite{Ding2014},\cite{Leimkuhler2016}). In general, the noise is neither Gaussian, spatially constant or isotropic. Nevertheless, there is numerical evidence that the thermostat increases stability and mixing. Heuristically, the dynamic for $\xi$ can be motivated by the fact that at equilibrium we have 
\begin{align*}
\mathbb{E}\left[ \frac{\partial^2 K}{\partial p_i^2}\right] = \int \frac{\partial^2 K}{\partial p_i^2} e^{-K(p)} dp =- \int \frac{\partial K}{\partial p_i}\left(-\frac{\partial K}{\partial p_i} e^{-K(p)}\right) dp
= \mathbb{E}\left[\left(\frac{\partial K}{\partial p_i}\right)^2\right]
\end{align*}
and hence $\mathbb{E}\left[\frac{d\xi}{dt}\right]=0$.
The additional dynamics pushes the system towards  $\frac{d\xi}{dt}=0$ suggesting that the distribution will be moved closer to the equilibrium. This gives a recipe for a stochastic gradient  Nosé-Hoover thermostat with a general kinetic energy $K(p)$.

We first augment the Hamiltonian with $\xi$:
\begin{align*}
	&H(q,p,\xi)=U(q) + K(p) + \frac{d}{2}(\xi-D)^2.
\end{align*}
We are now in the position to derive the SDE preserving the probability density $\propto \exp(-H)$ by adopting the framework of \cite{Ma2015} and defining:
\begin{align}
	H(\theta,p,\xi) &= U(\theta) + K(p) + \frac{d}{2}(\xi -D)^2\\
	D(\theta,p,\xi) &= \left(\begin{array}{ccc}
		0 & 0 & 0\\
		0 & D\cdot I & 0\\
		0 & 0 & 0
	\end{array}\right) \\
	Q(\theta,p,\xi) &= \left(\begin{array}{ccc}
		0 & -I & 0\\
		I & 0 & \nabla K(p)/d\\
		0 & -\nabla K(p)^T/d & 0
	\end{array}\right).
\end{align}
From (\ref{eq:Ma}) it follows that  $\Gamma = \left(0 \; 0\; -\Delta K(p)/d\right)^T$ and the dynamics becomes

\begin{center}
$\begin{pmatrix} 
d\theta \\ d p \\ d\xi
\end{pmatrix}  = 
\begin{pmatrix} 
 \nabla K(p) \\ -\nabla \tilde{U} dt - \xi \nabla K(p) \\
\frac{1}{d}\left(\|\nabla K(p)\|^2 - \Delta K(p)\right)
\end{pmatrix}  dt + 
\begin{pmatrix} 
0 & 0 & 0 \\ 
0 & \sqrt{2D} & 0 \\ 
0 & 0 & 0
\end{pmatrix} dW_t
$
\label{eq:genthermo}
\end{center}

where $\Delta$ is the Laplace operator defined as $\Delta K(p)= \sum_i \frac{\partial^2 K(p)}{\partial p_i^2}$. For the relativistic kinetic energy $K(p)$, we have that $\nabla_pK(p) = M^{-1}(p)p$ with $M(p):= m\left(\frac{p^Tp}{m^2c^2}+1\right)^\frac{1}{2}$ a scalar and that $ \Delta K(p) = tr\left(\frac{d}{dp}\left(\frac{1}{d}M^{-1}(p)p\right)\right)$. The Stochastic Gradient Nosé-Hoover Thermostat for relativistic HMC follows:

\begin{center}
$\begin{pmatrix} 
d\theta \\ d p \\ d\xi
\end{pmatrix}  = 
\begin{pmatrix} 
 M^{-1}(p)p  dt \\
- \nabla\tilde{U}-\xi M^{-1}(p)p  \\
\frac{p^Tp}{d}\left(M^{-2}(p) + c^{-2}M^{-3}(p)\right) - M^{-1}(p)
\end{pmatrix}  dt + 
\begin{pmatrix} 
0 & 0 & 0 \\ 
0 & \sqrt{2D} & 0 \\ 
0 & 0 & 0
\end{pmatrix} dW_t
$
\end{center}

%% file: toy_example.tex
\begin{figure}[t]
\centering
\begin{subfigure}{.45\textwidth}
  \centering
  \includegraphics[width=\linewidth]{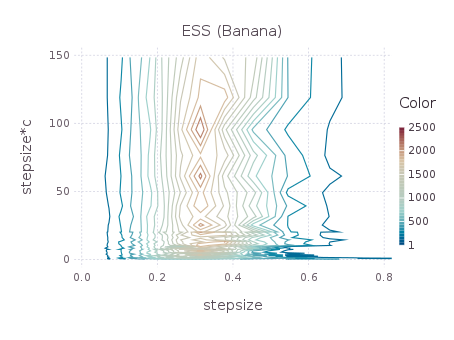}
\end{subfigure}%
\begin{subfigure}{.45\textwidth}
  \centering
  \includegraphics[width=\linewidth]{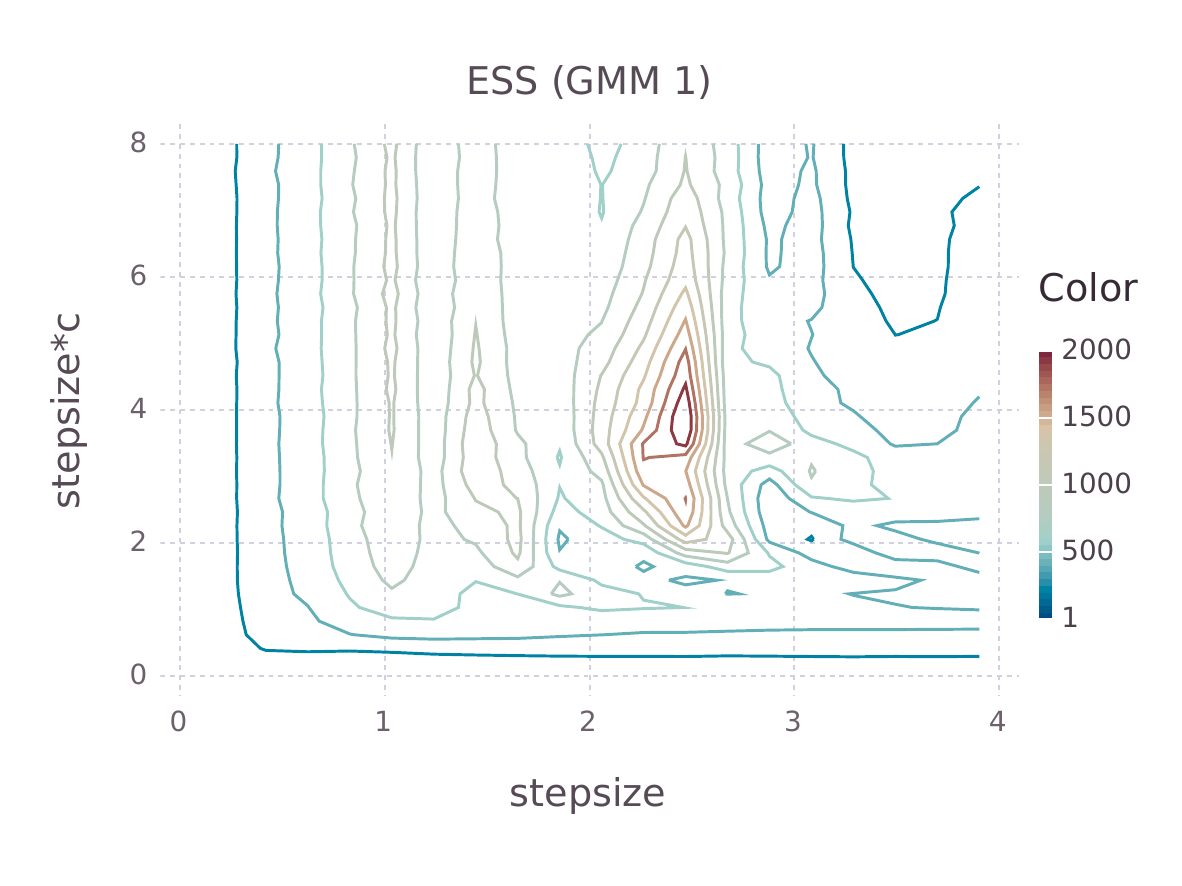}
\end{subfigure}
\vspace{-7pt}
\caption{ESS contour plots of $\epsilon \times c$ versus $\epsilon$ for  Banana  (left) and GMM1 (right) datasets.}
\label{fig:contour}
\end{figure}

\begin{figure}[t]
\centering
\begin{subfigure}{.3\textwidth}
  \centering
  \includegraphics[width=\linewidth]{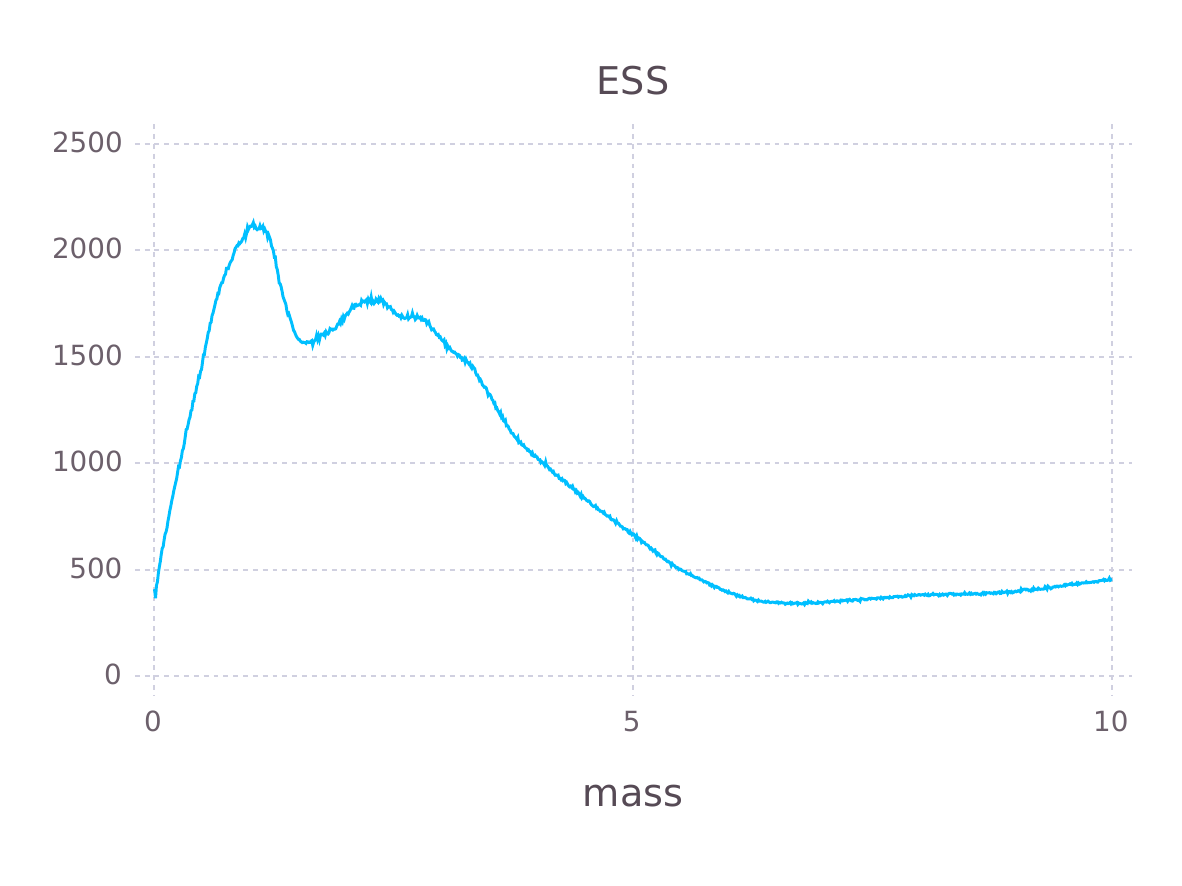}
\end{subfigure}
\begin{subfigure}{.3\textwidth}
  \centering
  \includegraphics[width=\linewidth]{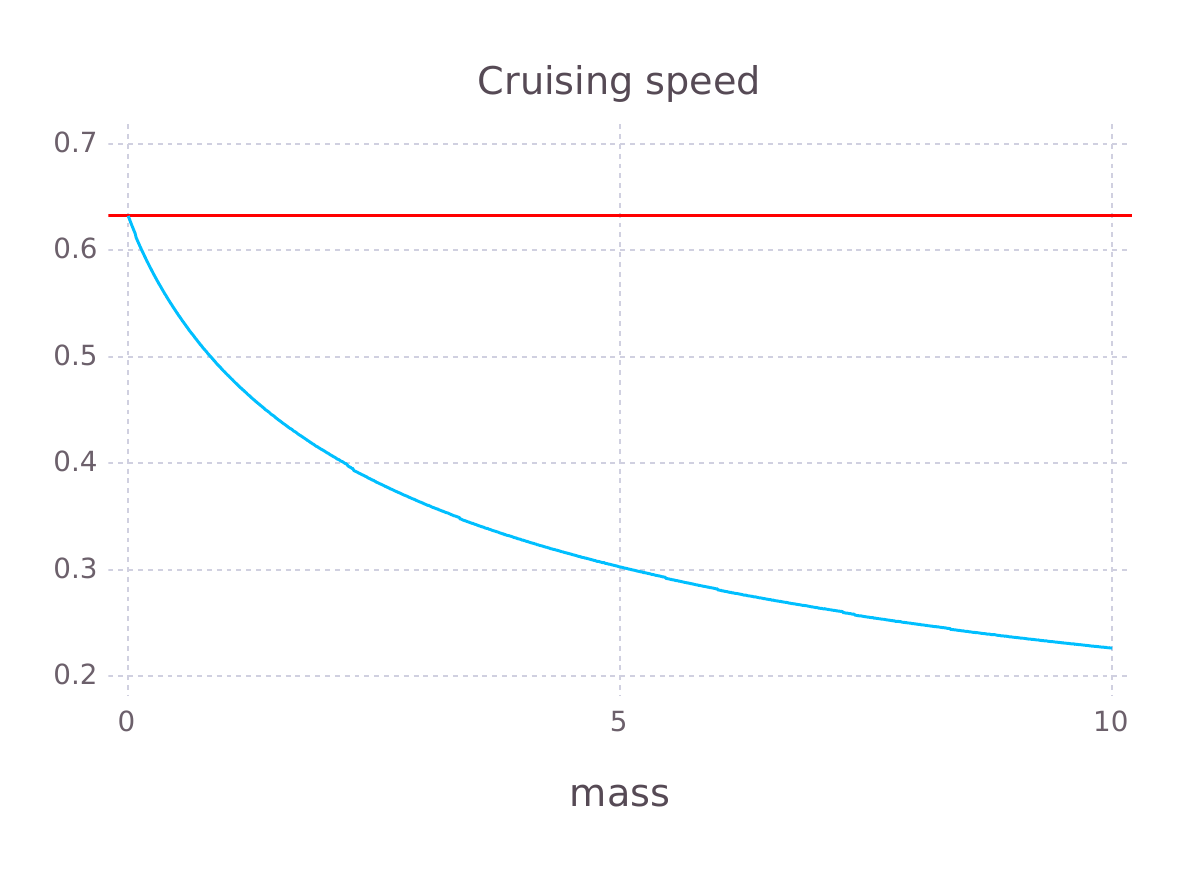}
\end{subfigure}%
\begin{subfigure}{.39\textwidth}
  \centering
  \includegraphics[width=\linewidth]{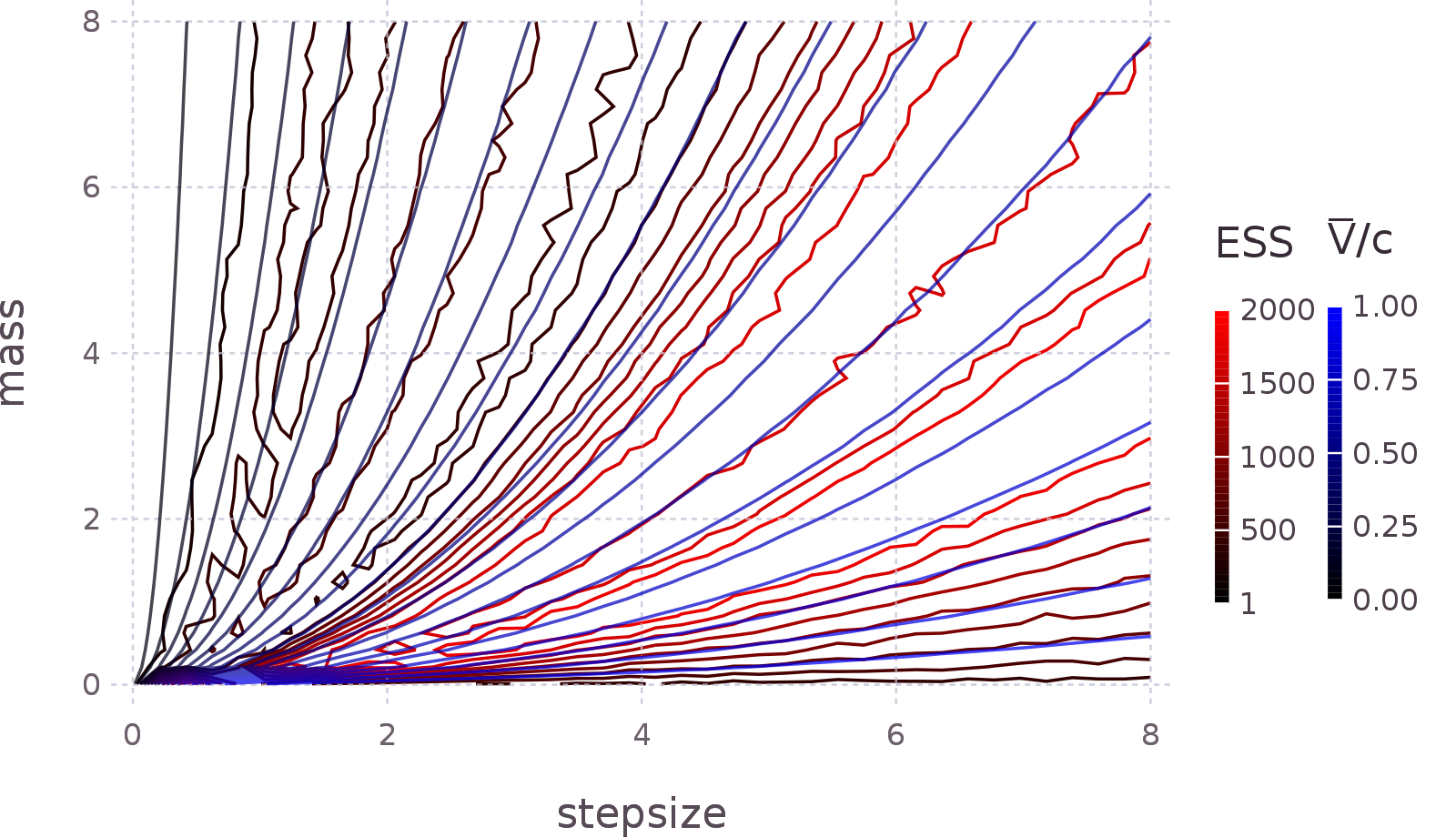}
\end{subfigure}%
\vspace{-7pt}
\caption{Varying $m$ for GMM1.  From left to right: ESS, cruising speed (the red horizontal line is $c$), and ESS and relative cruising speed $\bar{v}/c$ contour plots versus $m$ and $\epsilon$.}
\label{fig:gmm_mass}
\end{figure}
\begin{figure*}[t]
\centering
\begin{subfigure}{.24\textwidth}
  \centering
  \includegraphics[width=\linewidth]{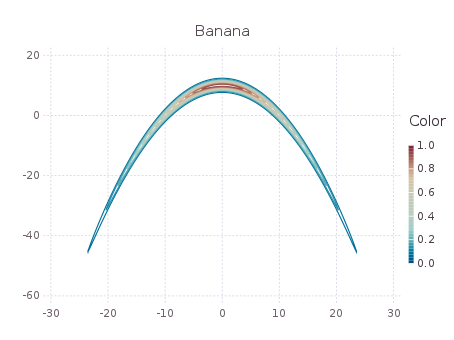}
\end{subfigure}%
\begin{subfigure}{.24\textwidth}
  \centering
  \includegraphics[width=\linewidth]{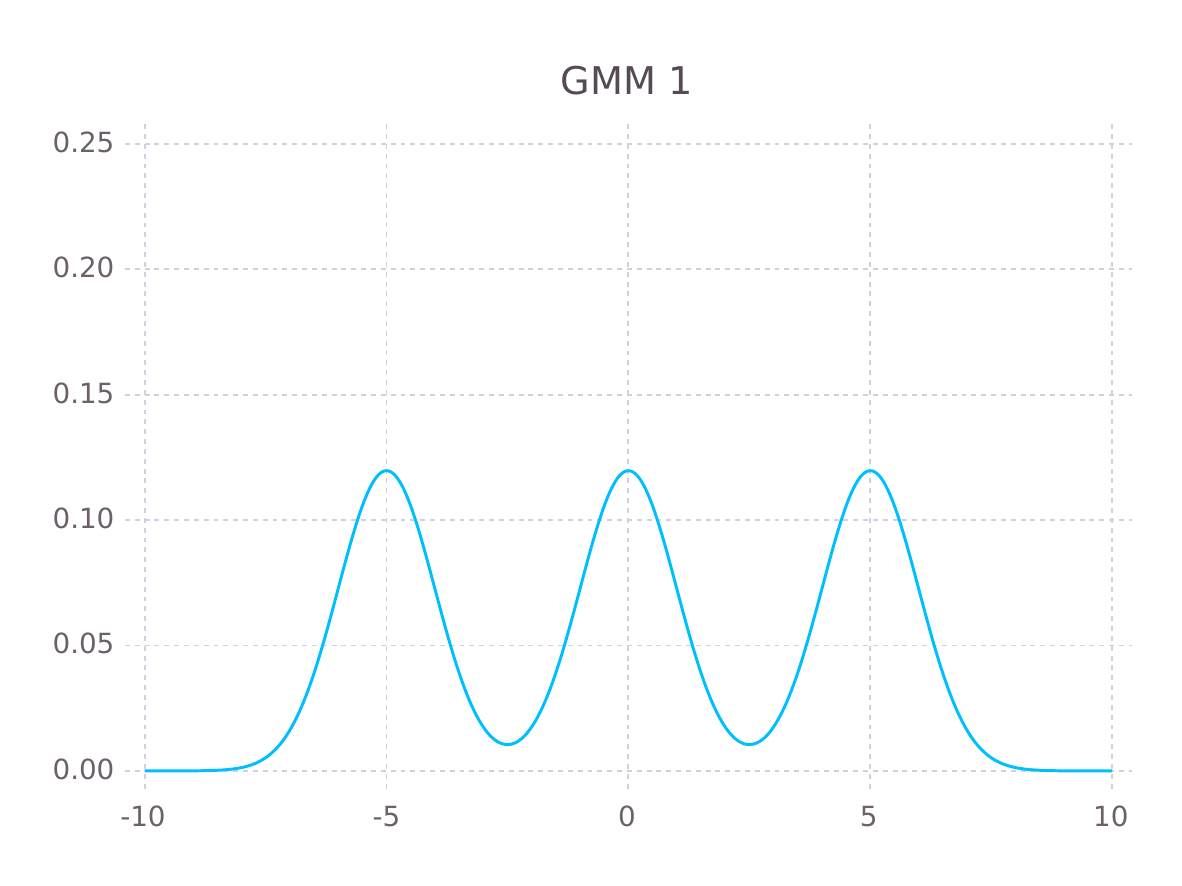}
\end{subfigure}
\begin{subfigure}{.24\textwidth}
  \centering
  \includegraphics[width=\linewidth]{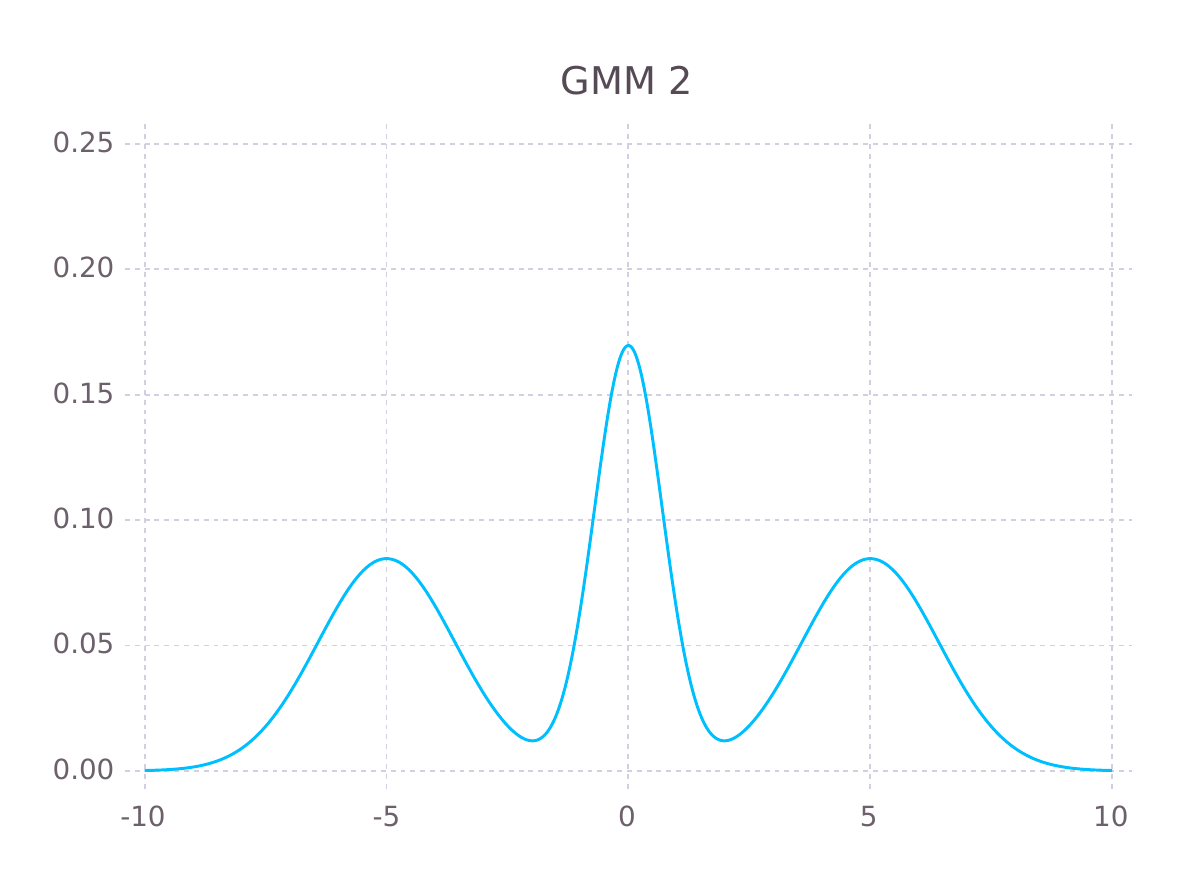}
\end{subfigure}
\begin{subfigure}{.24\textwidth}
  \centering
  \includegraphics[width=\linewidth]{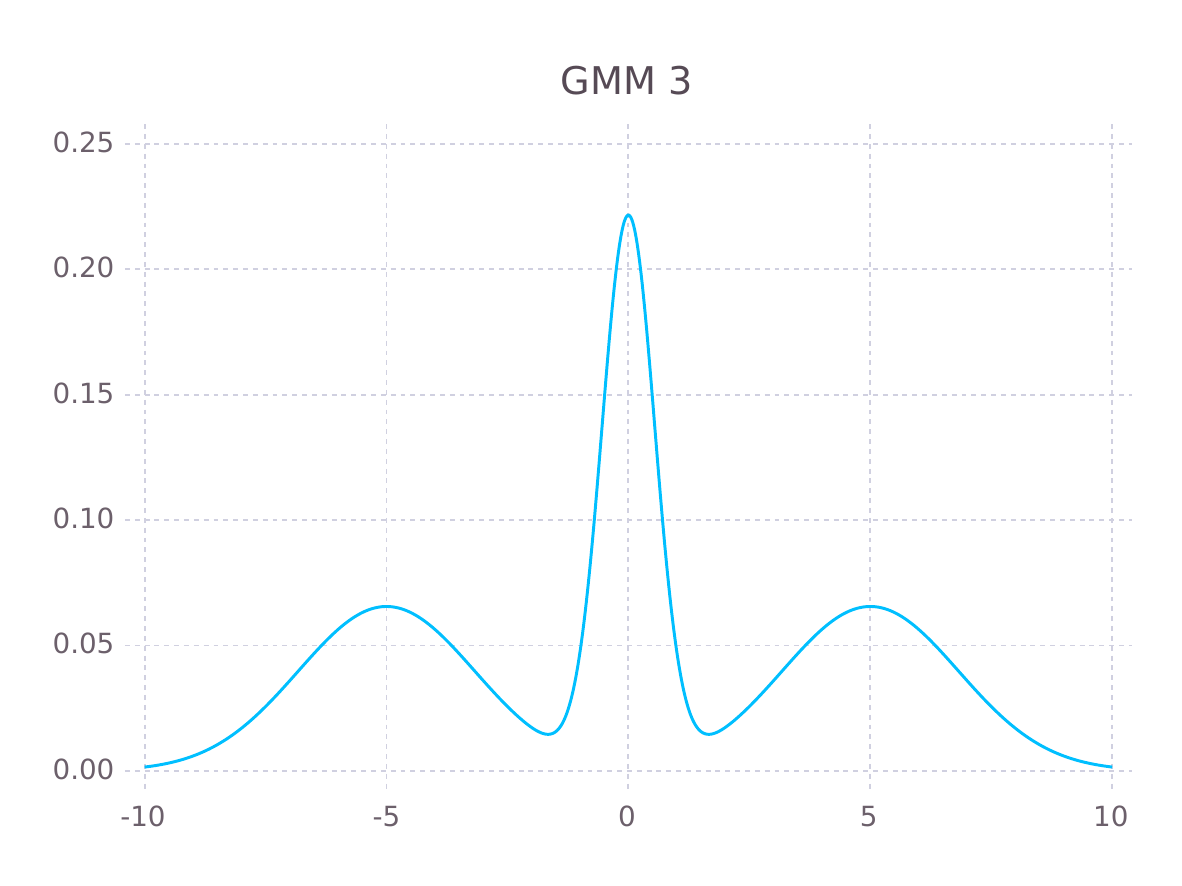}
\end{subfigure} \\
\vspace{-3mm}
\begin{subfigure}{.24\textwidth}
  \centering
  \includegraphics[width=\linewidth]{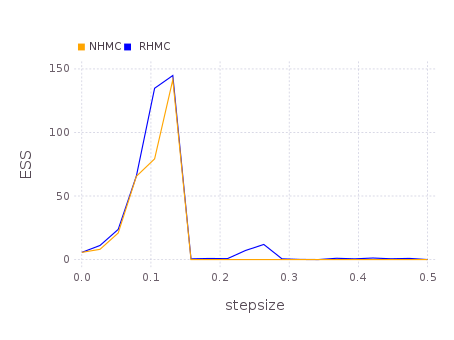}
\end{subfigure}%
\begin{subfigure}{.24\textwidth}
  \centering
  \includegraphics[width=\linewidth]{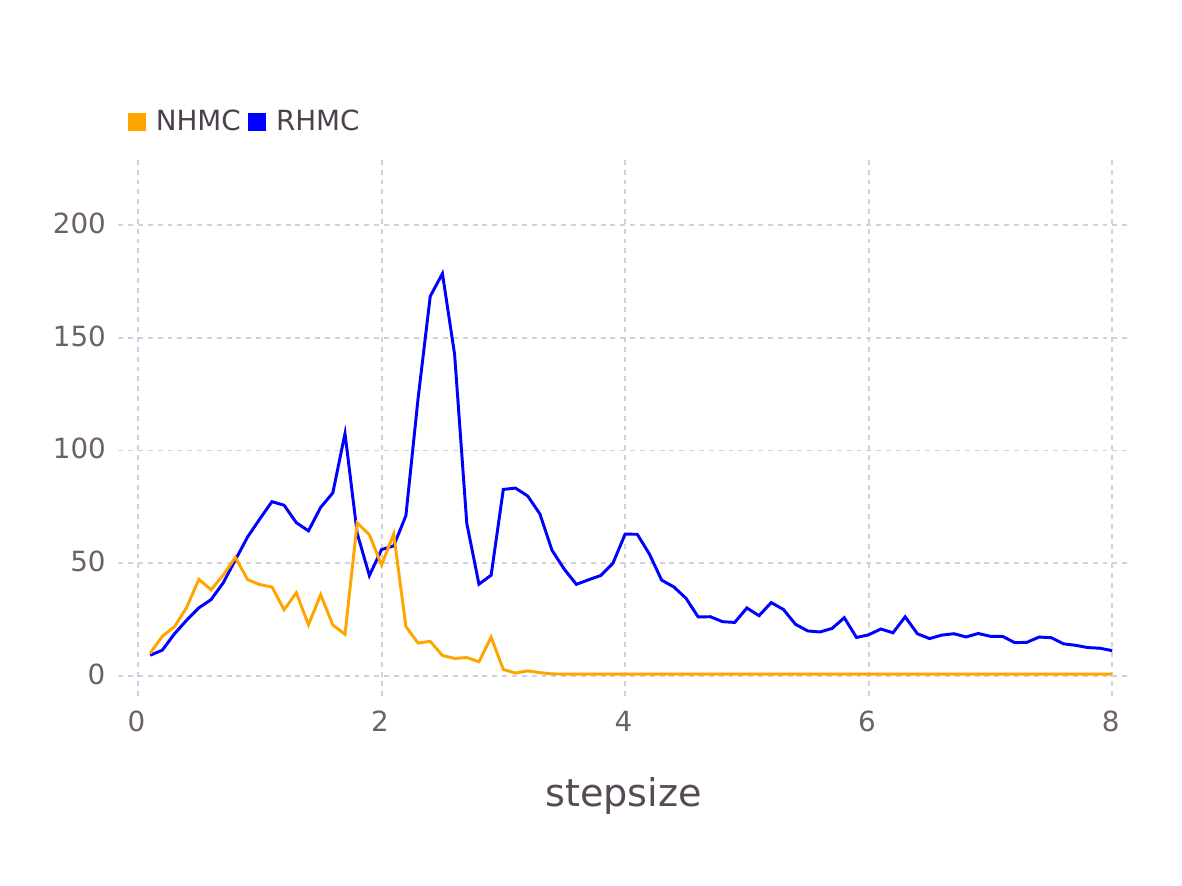}
\end{subfigure}
\begin{subfigure}{.24\textwidth}
  \centering
  \includegraphics[width=\linewidth]{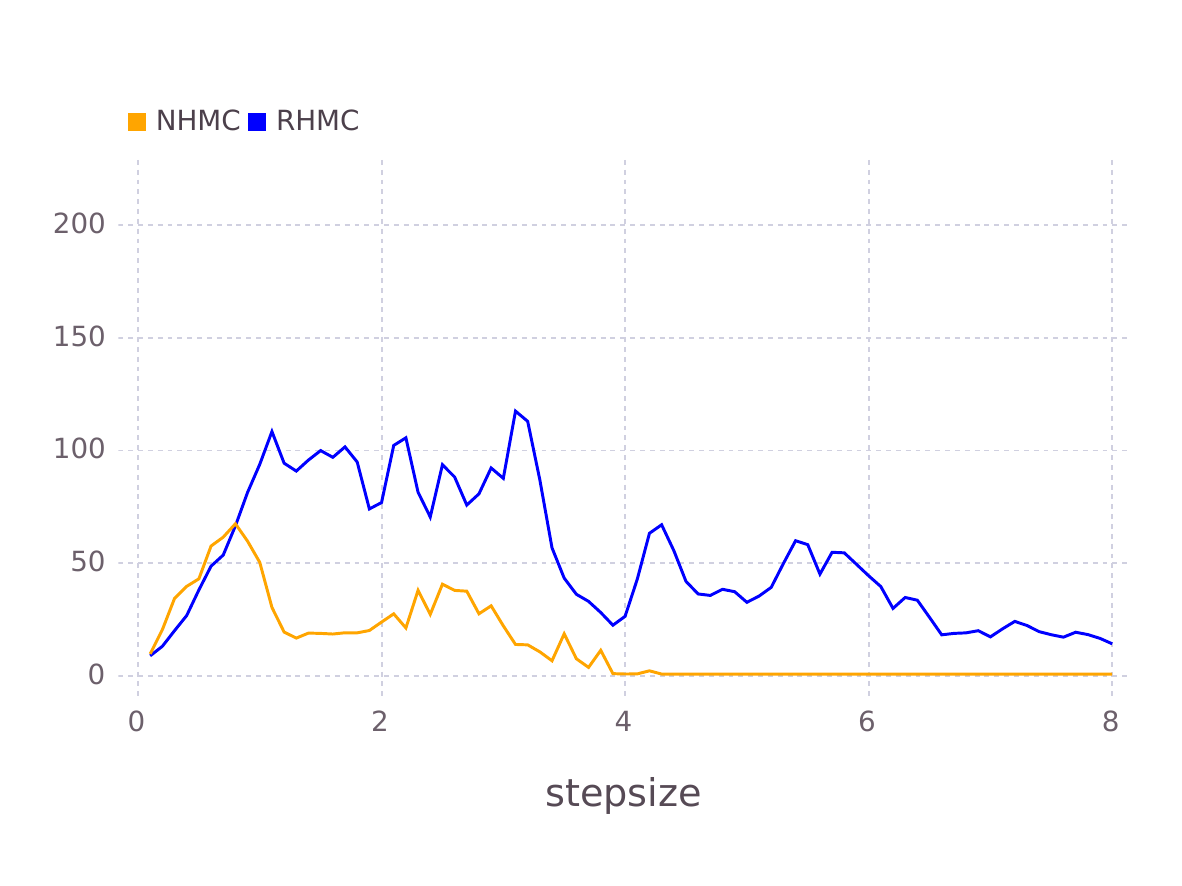}
\end{subfigure}
\begin{subfigure}{.24\textwidth}
  \centering
  \includegraphics[width=\linewidth]{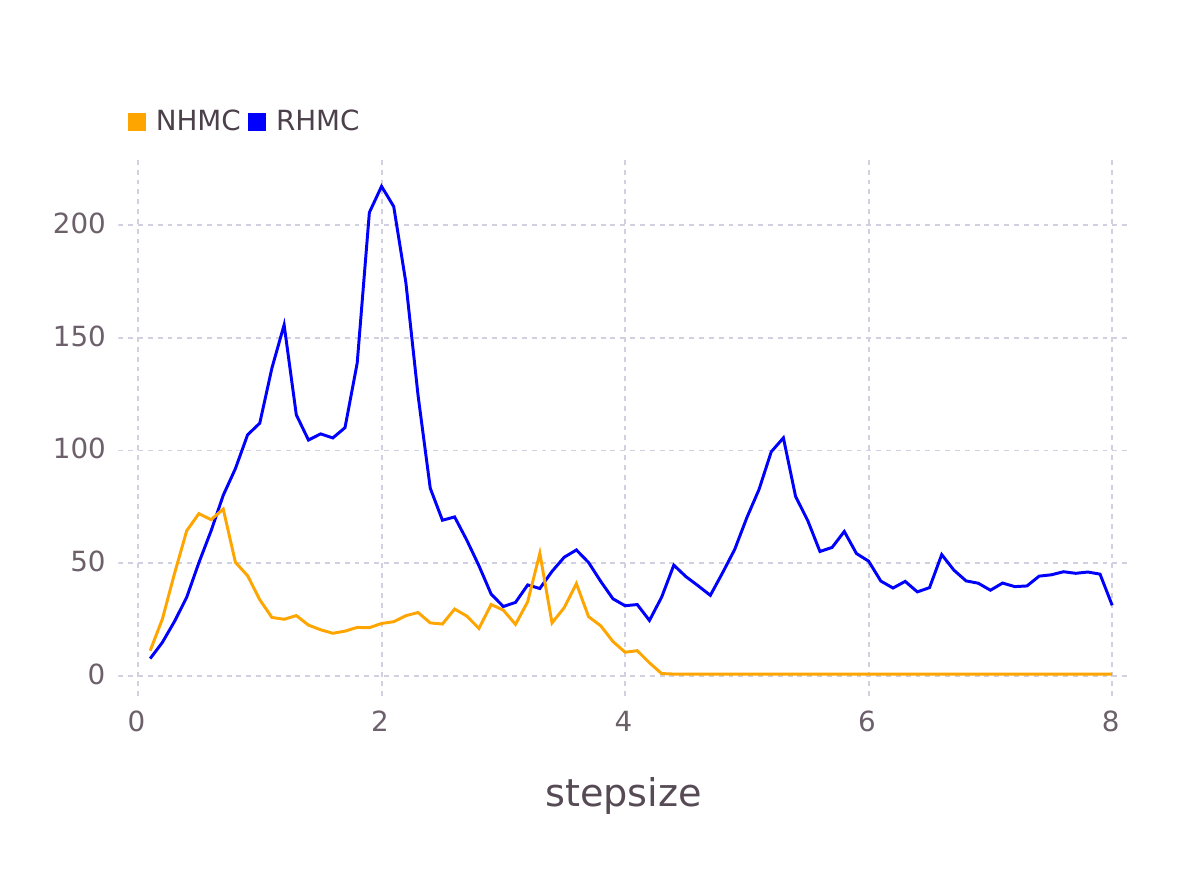}
\end{subfigure} \\
\vspace{-3mm}
\begin{subfigure}{.24\textwidth}
  \centering
  \includegraphics[width=\linewidth]{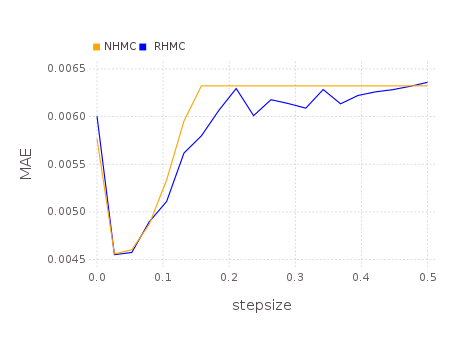}
\end{subfigure}%
\begin{subfigure}{.24\textwidth}
  \centering
  \includegraphics[width=\linewidth]{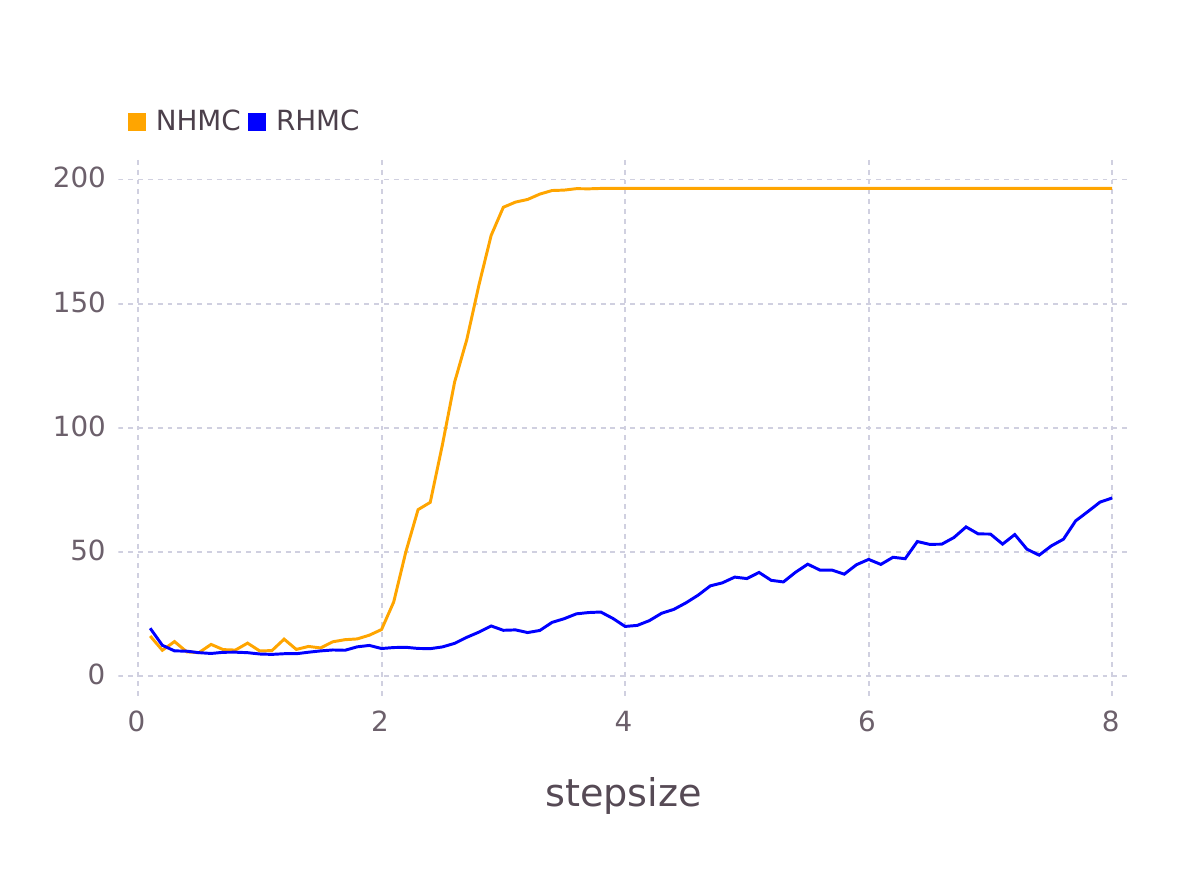}
\end{subfigure}
\begin{subfigure}{.24\textwidth}
  \centering
  \includegraphics[width=\linewidth]{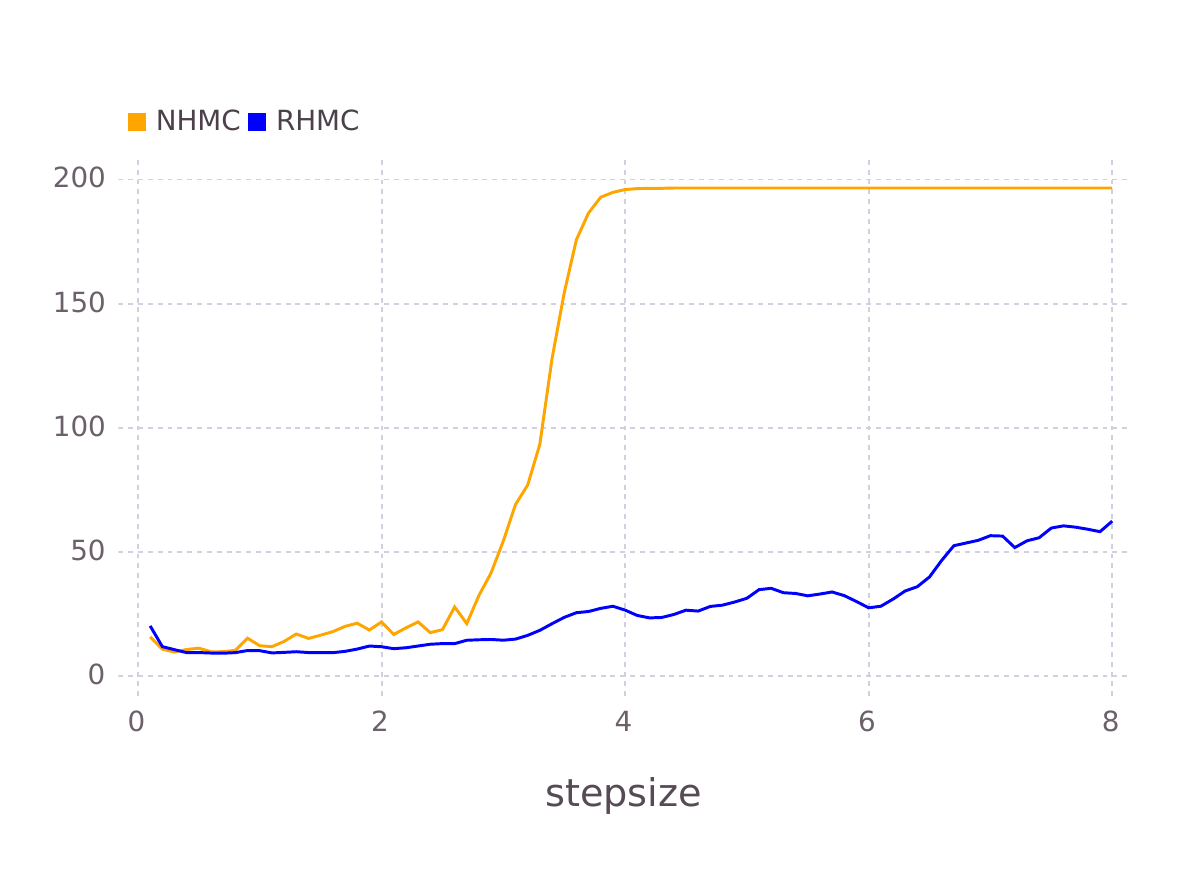}
\end{subfigure}
\begin{subfigure}{.24\textwidth}
  \centering
  \includegraphics[width=\linewidth]{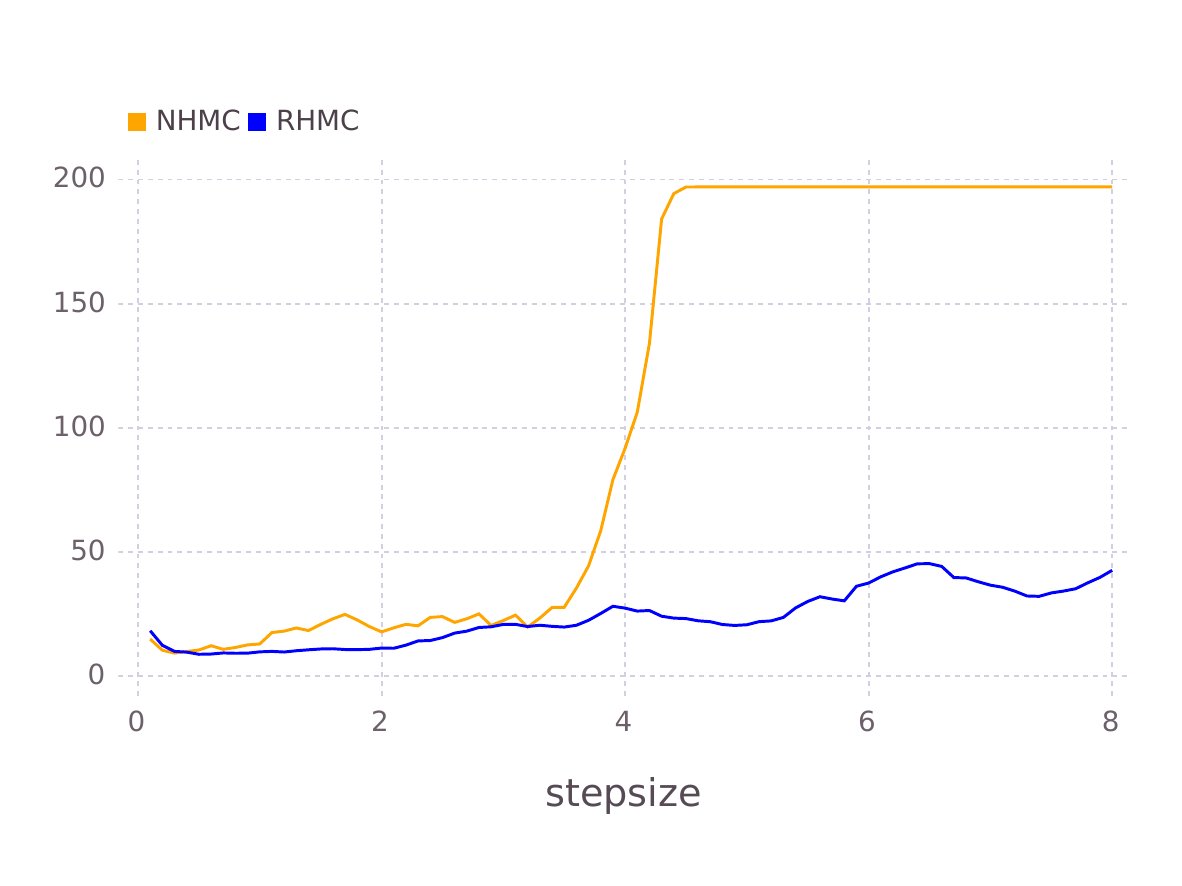}
\end{subfigure} \\
\vspace{-3mm}
\begin{subfigure}{.24\textwidth}
  \centering
  \includegraphics[width=\linewidth]{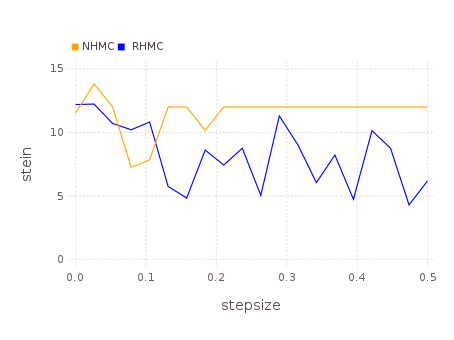}
\end{subfigure}%
\begin{subfigure}{.24\textwidth}
  \centering
  \includegraphics[width=\linewidth]{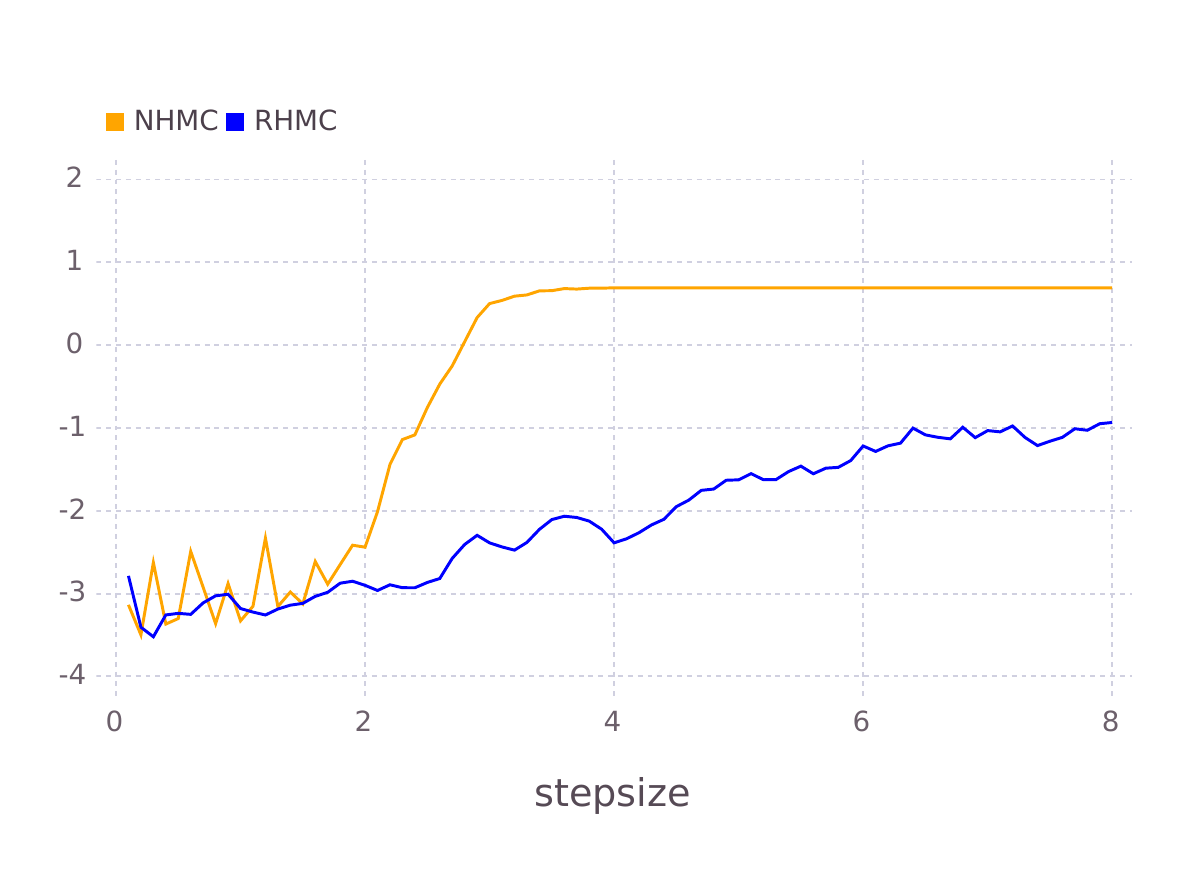}
\end{subfigure}
\begin{subfigure}{.24\textwidth}
  \centering
  \includegraphics[width=\linewidth]{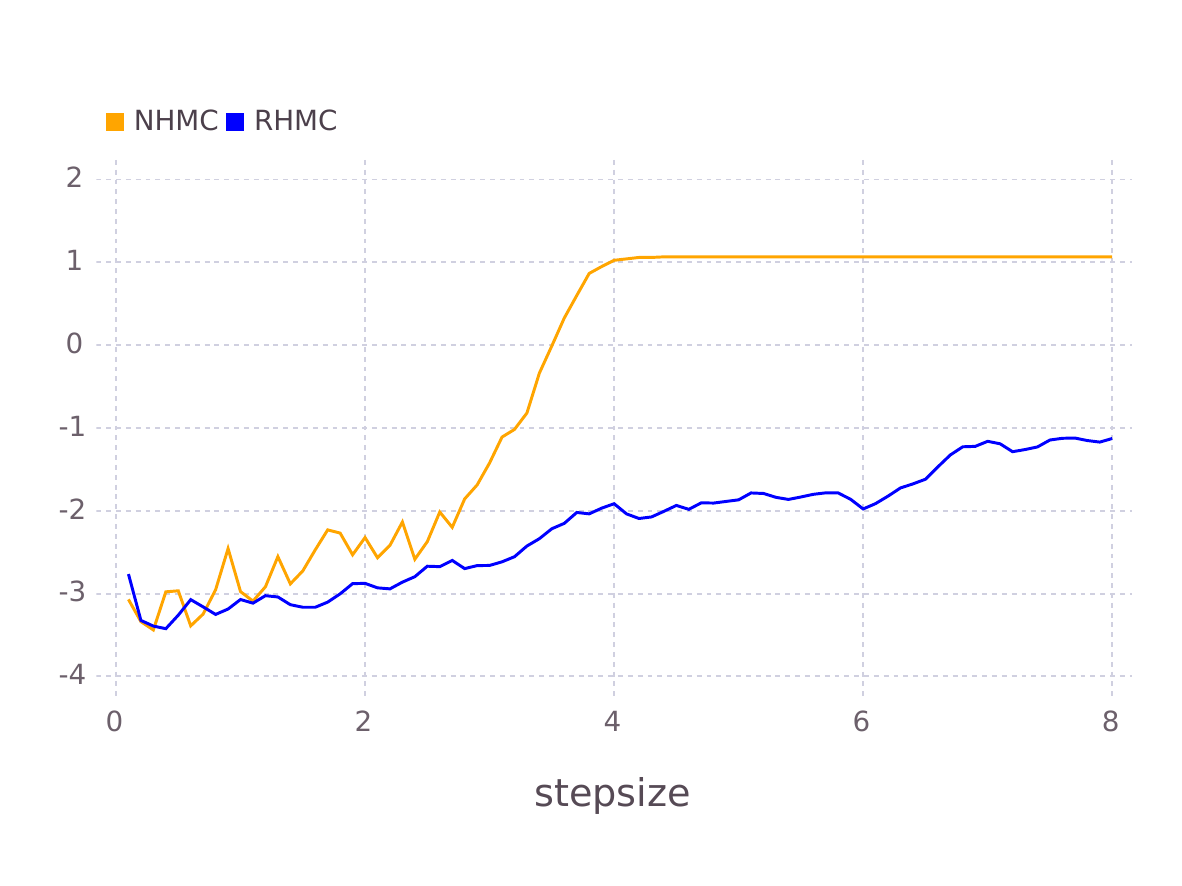}
\end{subfigure}
\begin{subfigure}{.24\textwidth}
  \centering
  \includegraphics[width=\linewidth]{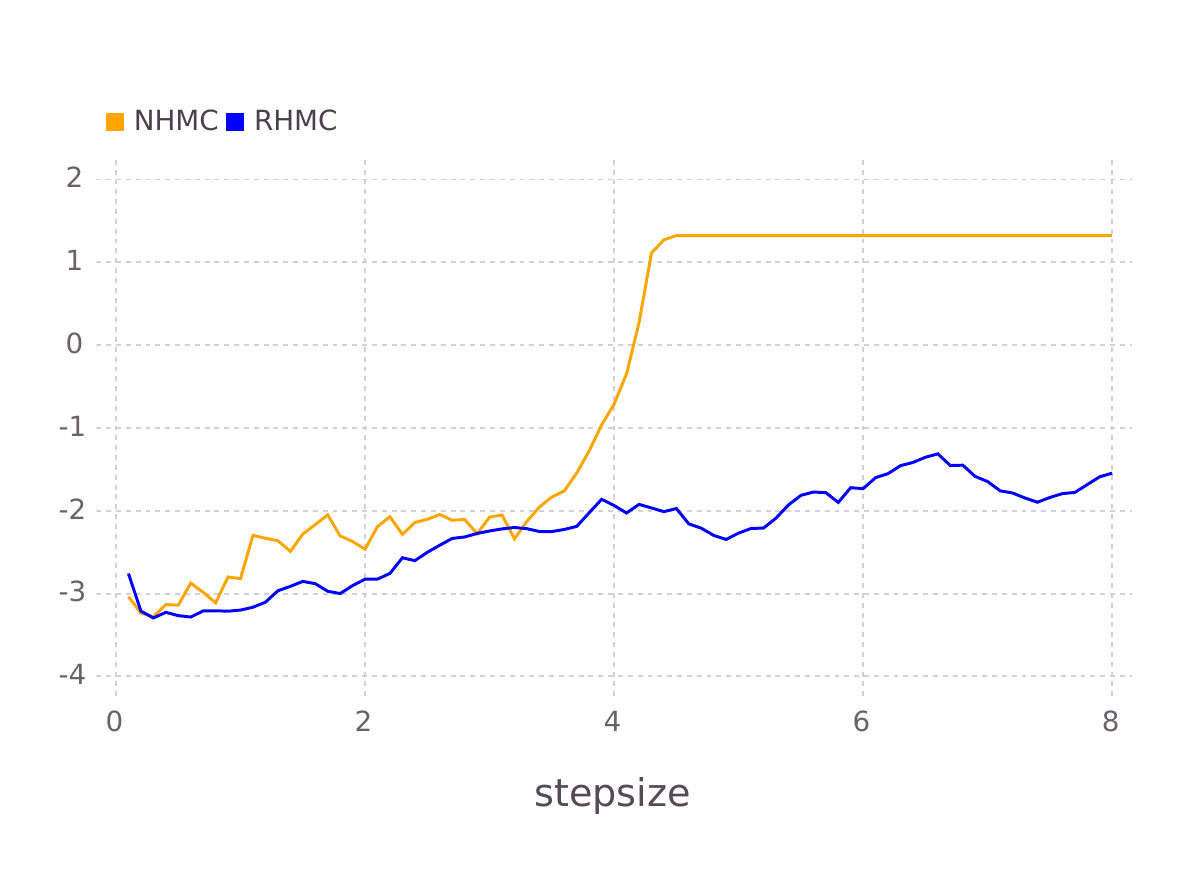}
\end{subfigure} 
\vspace{-10pt}
\caption{Left to right: Banana, GMM1, GMM2, GMM3 datasets. Top to bottom: density plot, ESS versus step size $\epsilon$, MAE versus $\epsilon$, log stein discrepancy versus $\epsilon$.}
\label{fig:toy}
\vspace{-20pt}
\end{figure*}

All the experimental results in this section are based on the separable versions (\ref{eq:separable}) as they give superior results than the non-separable counterparts.
We first explore the performances of the algorithms on a set of small examples including a two dimensional banana function (Banana) \cite{haario1999} with density $p(\mathbf{x}) \propto \exp\{-0.5(0.01x_1^2+(x_2+0.1x_1^2-10)^2)\}$, and Gaussian mixture models (GMM1, GMM2, GMM3) obtained by combining the three following Gaussian random variables with equal mixing proportions: $ \mathcal{N}(-5,1/ \sigma^2)$,  $\mathcal{N}(0,\sigma^2)$,  $\mathcal{N}(5,1/\sigma^2)$, where $\sigma^2 = 1, 0.5, 0.3$.  When $\sigma^2=1$ the three Gaussians have the same variance and lower $\sigma^2$ means larger the discrepancies between their variances and thus a wider range of length scales and log density gradients.  The density plots of the examples can be found in the top row of Figure \ref{fig:toy}. 

We start with an exploration of the behaviour of RHMC as the tuning parameters $m$, $c$ and $\epsilon$ are varied.  First we considered the effective sample sizes (ESS) of the algorithm on the Banana and GMM1 datasets.  We varied both $\epsilon$ and $\epsilon\times c$ over a grid, and computed the average ESS, over $20$ chains, each of length $10^4$ for Banana, and over 100 chains of length $10^5$ for GMM1. The ESS contour plots can be found in Figure \ref{fig:contour}, which suggests that $\epsilon c$ and $\epsilon$ can be independently tuned.  While $\epsilon$ controls the time discretisation of the continuous-time dynamics, $\epsilon c$ controls the maximum change in the parameters at each leapfrog step.  Next we varied the mass parameter $m$ for GMM1, showing plots in Figure \ref{fig:gmm_mass}.  As expected the ESS is optimised at an intermediate value of $m$, and the average ``cruising speed'' $\bar{v}$ decreases with $m$.  In order to understand how to tune $m$, on the fourth panel we overlaid two contour plots: one for ESS and the other for $\bar{v}$.  We see that the cruising speed $\bar{v}$ correlates much better with the ESS than $m$ does, which suggests that $m$ should be tuned via $\bar{v}$, e.g.\ by the user specifying a desired value for $\bar{v}$ and $m$ being adapted to achieve the speed (noting that $m$ and $\bar{v}$ have a monotonic relationship which makes for easy adaptation).

We next compare the performances of NHMC and RHMC for a wide range of step sizes, via the ESS (higher better), the \emph{mean absolute error (MAE)} between the true probabilities and the histograms of the sample frequencies (lower better), and the log Stein discrepancy \cite{Stein2015} which is a more accurate measure of sample quality (lower better). The reason being the Wasserstein distance can be bounded in terms of the Stein discrepancy thus accounting for bias and insufficient exploration of the target.  The results can be found in rows 2-4 of Figure \ref{fig:toy}. It can be seen that RHMC achieves better performance and is strikingly more robust to the step size $\epsilon$ than NHMC. As expected, this behaviour is particularly pronounced when the step size is large. Moreover, when the gradients of the target model span a large range of values (GMM2, GMM3), the improvements yielded by the relativistic variants are more pronounced. These results confirm that, since the speed of particles is bounded by $c$, RHMC is less sensitive to the presence of large gradients in the target density and more stable with respect to the choice of $\epsilon$, allowing for a more efficient exploration of the target density.

%% file: LR.tex
\begin{wrapfigure}{R}{0.45\textwidth}
  \vspace{-30pt}
\centering
  \hspace*{-2em}
  \includegraphics[width=.51\textwidth]{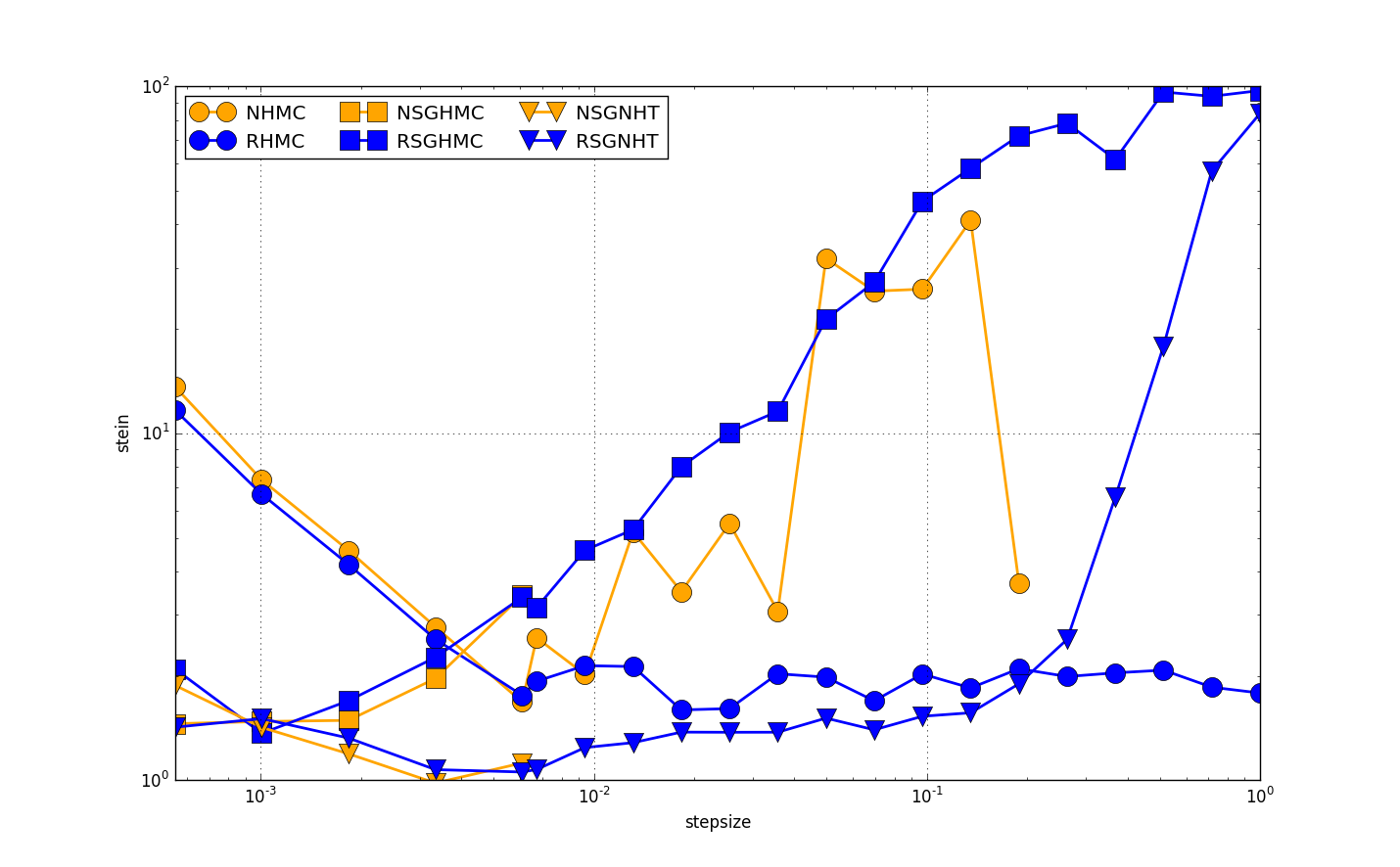}\hspace*{-2em}
    \vspace{-10pt}
  \caption{Stein discrepancy versus step size $\epsilon$ for logistic regression. NSGHMC and NSGNHT were unstable for $\epsilon>6\times 10^{-3}$ and thus their stein discrepancies were not plotted.}
    \label{fig:LR_stein}
    \vspace{-5pt}
\end{wrapfigure}

Next we compare both the Newtonian and relativisitic variants of HMC and SGMCMC algorithms on a simulated 3-dimensional logistic regression example with $500$ observations. For the stochastic versions of the algorithms, we use mini-batches of size $100$. After a burn-in period of $1000$ iterations, we calculated the Stein discrepancy for different $\epsilon$ while keeping the product $\epsilon \times c$ fixed. To make a fair comparison, we used $200$ samples for NHMC and RHMC and $1000$ samples for the SGMCMC algorithms. From Figure \ref{fig:LR_stein}, we see that the relativistic variants are significantly more robust than the Newtonian variants.  The NHT algorithms were able to correct for stochastic gradient noise and performed better than SGHMC algorithms.  Particularly, RSGNHT had lower Stein discrepencies than other algorithms for most values of $\epsilon$.

%% file: MNIST.tex
\begin{figure*}[h!]
\centering
\begin{subfigure}{.31\textwidth}
\centering
\includegraphics[width=\linewidth]{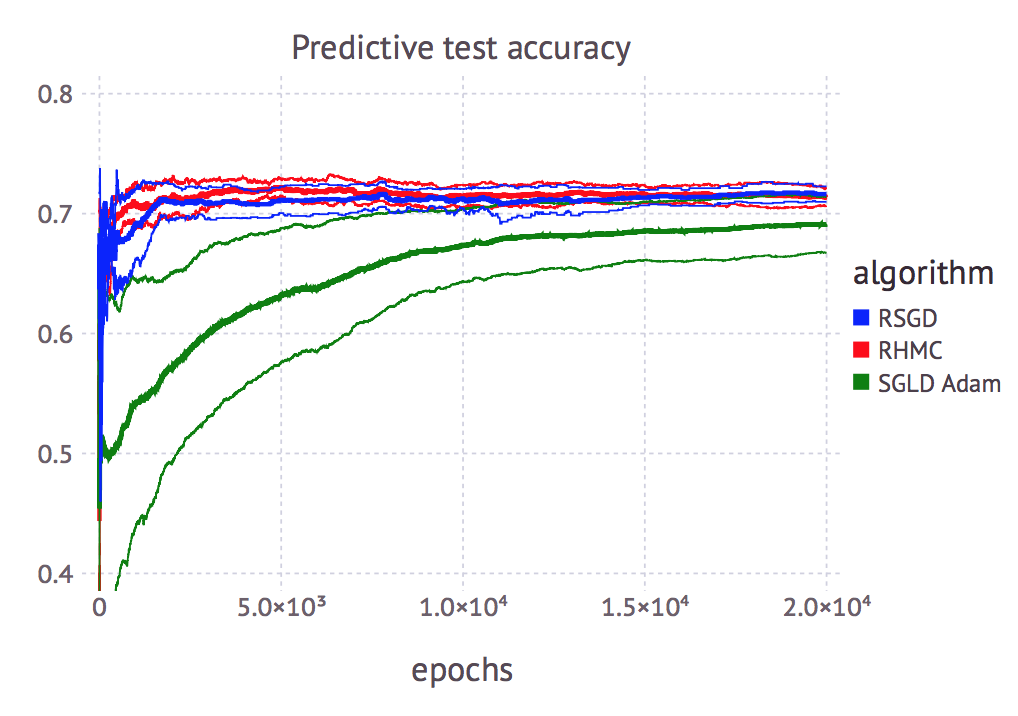}
\end{subfigure}
\begin{subfigure}{.22\textwidth}
\centering
\includegraphics[width=\linewidth]{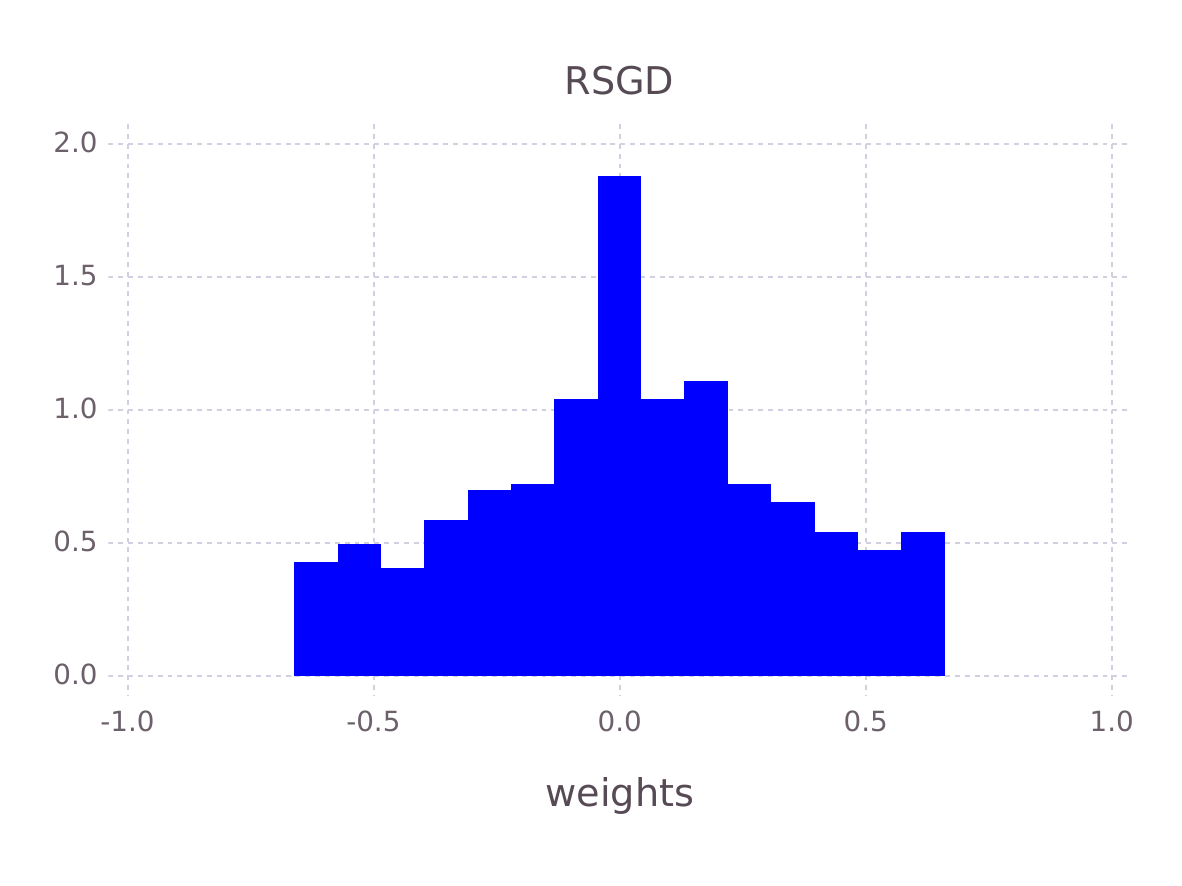}
\end{subfigure}
\begin{subfigure}{.22\textwidth}
\centering
\includegraphics[width=\linewidth]{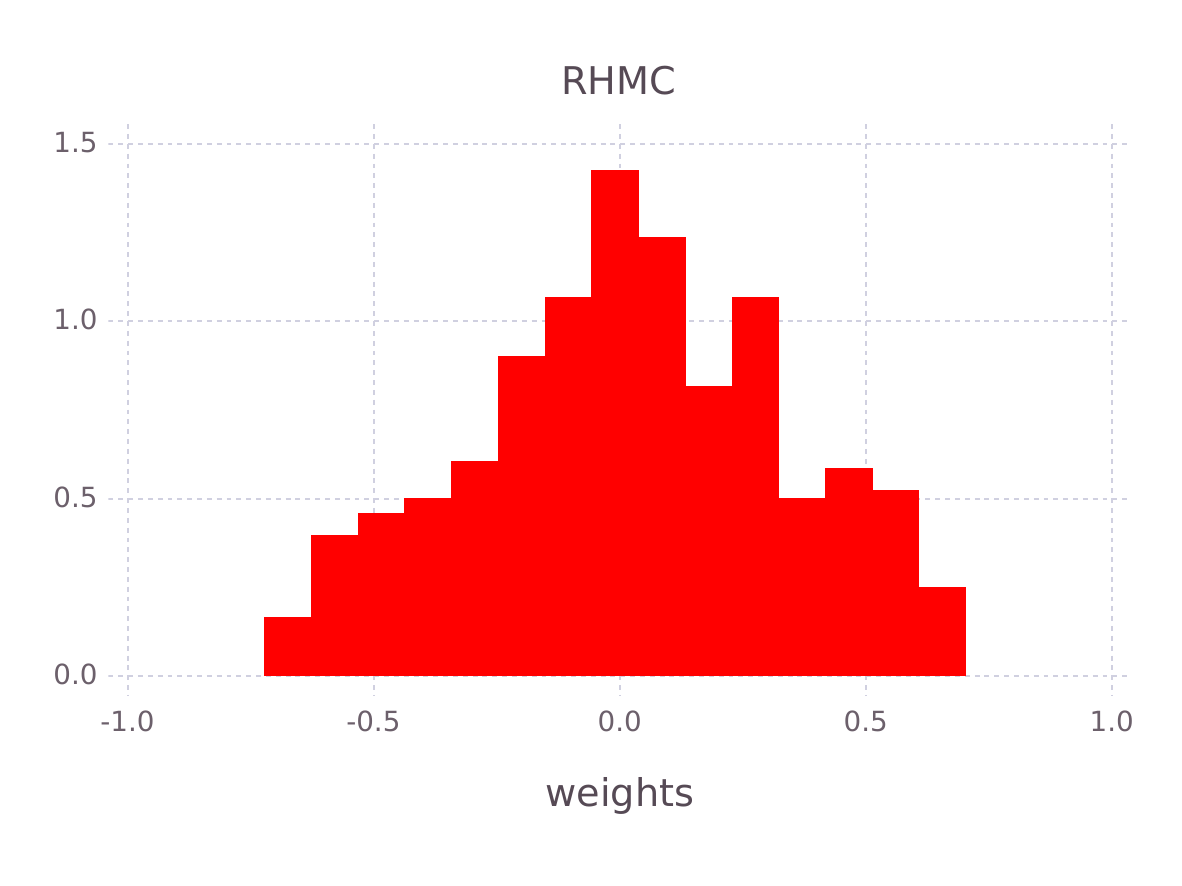}
\end{subfigure}
\begin{subfigure}{.22\textwidth}
\centering
\includegraphics[width=\linewidth]{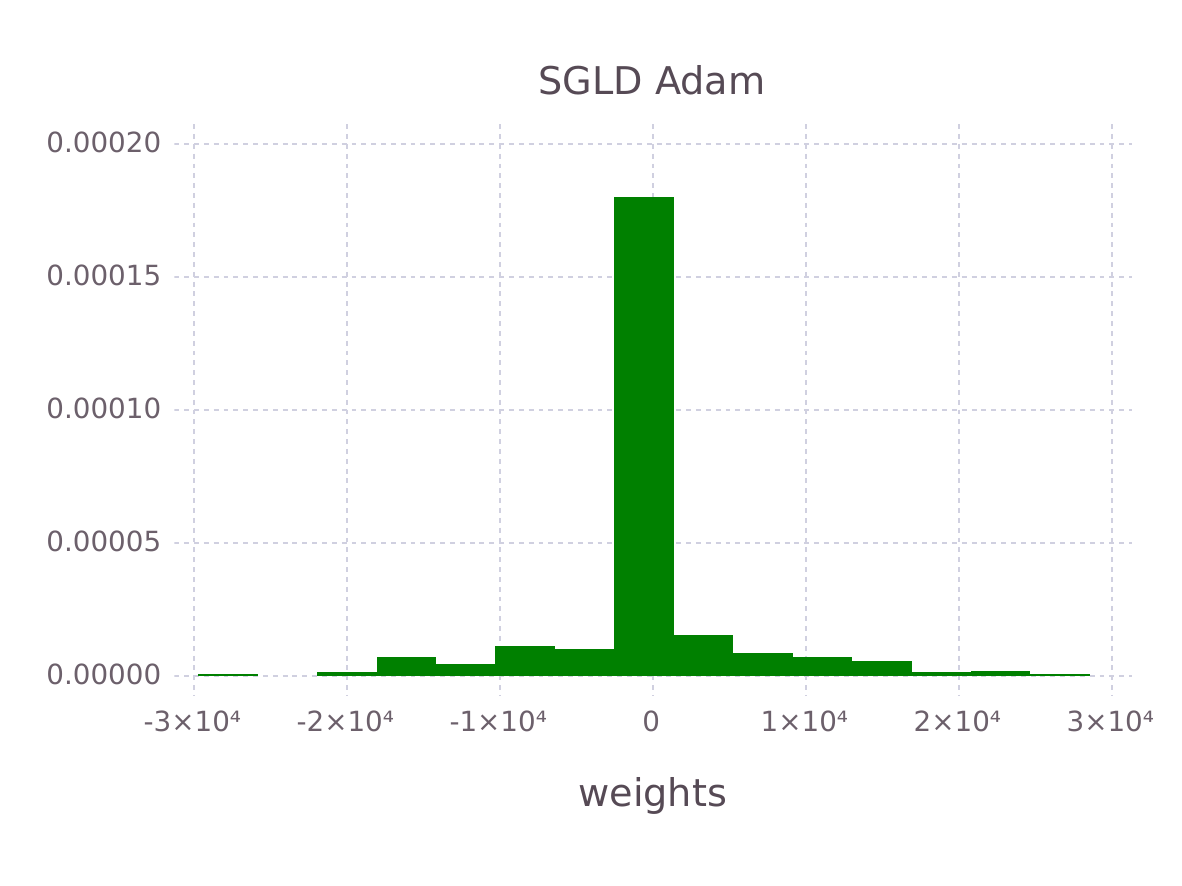}
\end{subfigure}

\caption{Comparison between RSGD, RHMC and SGLD Adam on the Pima Indians dataset using 50 hidden units. The histograms show the neural network weights at the first iteration.}
\label{fig:weights}
\end{figure*}

Turning to more complex models, we first considered a neural network with 50 hidden units and initialized its weights by the widely used Xavier initialization. We used the Pima Indians dataset for binary classification (552 observations and 8 covariates) to compare the relativistic and the preconditioning approach. Indeed, these methods represent two different ways to normalise gradients so that the update sizes are reasonable for the local lengthscale of the target distribution. In particular we consider SGLD Adam, namely a preconditioned SGLD algorithm with an additional Adam-style debiasing of the preconditioner. Figure \ref{fig:weights} compares the test-set accuracy of SGLD Adam with RSGD and RHMC, showing that the first is significantly outperformed by the relativistic algorithms. Due to Xavier initialization, all of the weights are small which causes small gradients, therefore the injected noise becomes very large due to the rescaling by the inverse of squared root of the average gradients, which makes SGLD Adam unstable. The histograms reveal that at the first iteration SGLD Adam causes the weights to become extremely large and this strongly compromises the performance of SGLD Adam, which takes a long time to recover. The relativistic framework represents therefore a much better approach to perform adaptation of the learning rates specific to each parameter.

\begin{figure}[!h]
\centering
\begin{subfigure}{.33\textwidth}
\centering
  \includegraphics[width=\linewidth]{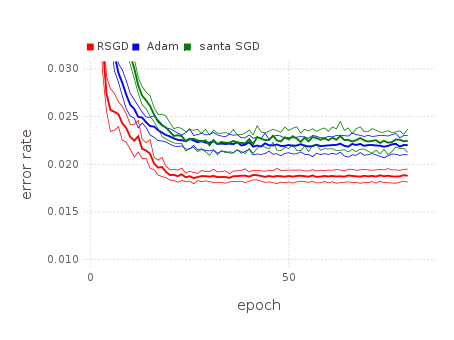}
\end{subfigure}%
\begin{subfigure}{.33\textwidth}
\centering
  \includegraphics[width=\linewidth]{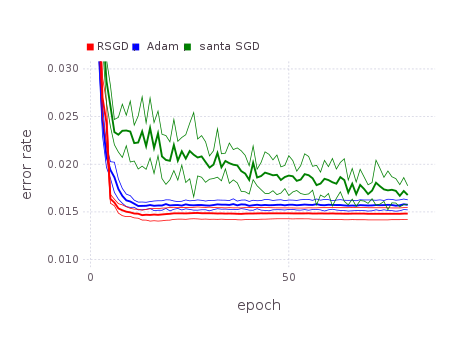}
\end{subfigure}
\begin{subfigure}{.33\textwidth}
\centering
  \includegraphics[width=\linewidth]{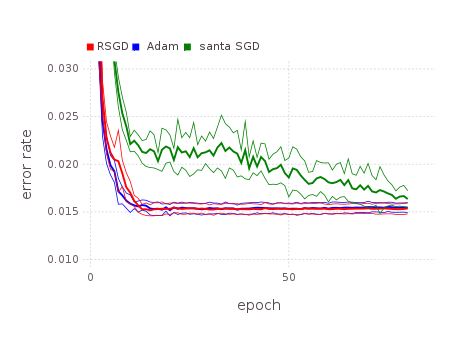}
\end{subfigure}
\vspace{-3mm}
\caption{Comparison of error rate on MNIST dataset on the test set. From left to right: $100$ hidden units; $500*300$ hidden units; $400*400$ hidden units.}
  \label{fig:MNIST}
\end{figure}

We then apply our algorithms to the standard MNIST dataset, which consists $28 \times 28$ handwritten digital images from 10 classes with a training set of size $60,000$ and a test set of size $10,000$. We tested our optimization algorithm on a single layer with $100$ hidden units and a multi-layer neural network with $500*300$ hidden units. In Figure \ref{fig:MNIST}  a comparison with Adam and Santa \cite{chen2015bridging} is displayed, their relation is discussed in more detail in Section \ref{sec:RSGD}. Note that, to ensure a fair comparison, we consider Santa SGD, namely a version of Santa that does not make use of symmetric splitting and simulated annealing. In other words, we adopt an Euler integration scheme for all algorithms and consider the zero-temperature limit for Santa. It can be observed that our algorithm is competitive with Adam and is able to achieve a lower error rate, particularly with the $100$ hidden units architecture. Moreover, RSGD performs significantly better than Santa SGD on all the considered architectures.

%% file: conclusion.tex
\section{Conclusion}

Our numerical experiments demonstrate that the relativistic algorithms discussed in this article are much more stable and robust to the choice of parameters and noise in stochastic gradients compared to the Newtonian counterparts. Moreover, we have a good understanding on how to choose  the  parameters $c$, $m$  and $\epsilon$. First the discretization parameter $\epsilon$ needs to be set, then we choose the maximal step $c\cdot \epsilon$ and in relation we choose the "cruising speed" $\frac{\bar{V}}{c}$ by picking $m$. The connection of our algorithms with popular stochastic optimizers such as Adam and RMSProp is novel and gives an interesting perspective to understand them.

Each of the proposed methodologies has scope for further research. The HMC version of the algorithm could be improved by employing some more advanced HMC methodology such as the NUTS version \cite{NUTS} and  using partial moment refreshment  instead of Adaptive Rejection Sampling \cite{Neal}. 
The relativistic stochastic gradient descent seems to be very competitive with state of the art stochastic gradient methods for fitting neural nets. Additionally, better numerical integration schemes could be employed. We also anticipate a variety of algorithms with different kinetic energies to be developed following our work. Last but not least, the strong simulation evidence should be complemented by more theoretical insights.

%% file: ack.tex
\section*{Acknowledgement}
XU thanks the PAG scholarschip and New College for support. LH and VP is funded by the EPSRC doctoral training centre OXWASP through EP/L016710/1.   YWT gratefully acknowledges EPSRC for research funding through grant EP/K009362/1.   SJV thanks EPSRC for funding through EPSRC Grants EP/N000188/1 and EP/K009850/1.